\documentclass[article]{elsarticle}
\usepackage{lineno,hyperref}

\usepackage{lineno,hyperref}
\usepackage{amsmath,amssymb, amsthm}
\usepackage{newtxtext,newtxmath}
\usepackage{graphicx}
\usepackage{latexsym}
\usepackage{color}
\usepackage{caption}
\usepackage{subcaption}
\usepackage{bm}
\usepackage{bbm}
\usepackage{algorithm}

\usepackage{algpseudocode}
\usepackage{color,soul}
\usepackage{graphics}
\usepackage{amsfonts}
\usepackage{setspace}   
\usepackage{wasysym}
\usepackage{multirow}
\usepackage{float}
\usepackage{standalone}
\usepackage{tikz}

\newcommand*\diff{\mathop{}\!\mathrm{d}}

\usepackage{epsfig} 

\makeatletter
\newcommand{\thickhline}{%
	\noalign {\ifnum 0=`}\fi \hrule height 1pt
	\futurelet \reserved@a \@xhline
}

\newif\ifboldnumber
\newcommand{\boldnext}{\global\boldnumbertrue}

\algrenewcommand\alglinenumber[1]{%
	\footnotesize\ifboldnumber\bfseries\fi\global\boldnumberfalse#1:}

\let\OldStatex\Statex
\renewcommand{\Statex}[1][3]{%
	\setlength\@tempdima{\algorithmicindent}%
	\OldStatex\hskip\dimexpr#1\@tempdima\relax}
\makeatother

\makeatletter
\def\ps@pprintTitle{%
	\let\@oddhead\@empty
	\let\@evenhead\@empty
	\def\@oddfoot{\reset@font\hfil\thepage\hfil}
	\let\@evenfoot\@oddfoot
}
\makeatother

\begin{document}
	
	\begin{frontmatter}
		
		\title{Adaptive Physics-Informed Neural Networks for~Markov-Chain Monte Carlo}
		
\author{Mohammad Amin Nabian$^{1,2}$}
\author{Hadi Meidani$^{1,*}$}
\address{$^1$ University of Illinois at Urbana-Champaign, Urbana, Illinois, USA. \\
	$^2$ NVIDIA, Santa Clara, California, USA. \\
	$^*$ Corresponding author (meidani@illinois.edu).}

		\begin{abstract}
		In this paper, we propose the Adaptive Physics-Informed Neural Networks (APINNs) for accurate and efficient simulation-free Bayesian parameter estimation via Markov-Chain Monte Carlo (MCMC). We specifically focus on a class of parameter estimation problems for which computing the likelihood function requires solving a PDE. The proposed method consists of: (1) constructing an offline PINN-UQ model as an approximation to the forward model; and (2) refining this approximate model on the fly using samples generated from the MCMC sampler. The proposed APINN method constantly refines this approximate model on the fly and guarantees that the approximation error is always less than a user-defined residual error threshold. We numerically demonstrate the performance of the proposed APINN method in solving a parameter estimation problem for a system governed by the Poisson equation.
		\end{abstract}

		
	\end{frontmatter}
	
	

\section{Introduction} 
In many engineering systems, there exists a set of parameters of interest that cannot be directly measured, and instead, one has to use indirect observations to estimate these parameters. This is  usually performed using the Bayes' theorem. Examples include estimating the parameters of a low-fidelity turbulence model, given direct or simulated measurements from a flow field, or estimating the reaction rate coefficients using measurements from the mass of components in a chemical reaction. 

Consider a parameter estimation problem for parameters $\mathbf{p} \in \mathbb{R}^{n_p}$ via Bayesian inference. Using the Bayes' theorem, the posterior density of parameters are obtained as follows

\begin{equation} \label{C8_posterior}
\pi\left ( \mathbf{p} | \mathbf{d} \right ) = \frac{\pi\left ( \mathbf{p} \right ) \pi \left ( \mathbf{d} | \mathbf{p} \right )}{\int \pi\left ( \mathbf{p} \right ) \pi \left ( \mathbf{d} | \mathbf{p} \right ) \diff \mathbf{p}},
\end{equation}
where $\mathbf{d}\in \mathbb{R}^{n_d}$ is the set of observations, $\pi\left ( \mathbf{p} | \mathbf{d} \right )$ is the posterior density of parameters, $\pi \left ( \mathbf{p}\right)$ specifies the prior density over the parameters, and $\pi \left(\mathbf{d} |  \mathbf{p} \right )$ is the likelihood function, which is built based on a deterministic forward model $f$ and a statistical model for the modeling error and measurement noise. Here we assume a zero modeling error and a zero-mean additive measurement noise, that is,

\begin{equation} \label{C8_error model}
\mathbf{d} = f(\mathbf{p})+\mathbf{\epsilon },
\end{equation}
where $\mathbf{\epsilon } \sim \pi_{\mathbf{\epsilon }} \left( \mathbf{\epsilon } \right)$. Therefore, the likelihood function can be represented as

\begin{equation} \label{C8_likelihood}
\pi \left ( \mathbf{d} | \mathbf{p} \right )=\pi_{\mathbf{\epsilon }}\left( f( \mathbf{p}) - \mathbf{d} \right ).
\end{equation}
Computing the posterior distribution in Equation \ref{C8_posterior} analytically often requires calculating intractable integrals. Alternatively, sampling-based methods, such as Markov Chain Monte Carlo (MCMC) methods \citep{brooks2011handbook,gelfand1990sampling,tierney1994markov,besag1995bayesian,chen2012monte,haario2006dram,dunson2020hastings,robert2020markov} may be used, where we use Markov chains to simulate samples for estimating the posterior distribution $\pi\left ( \mathbf{p} | \mathbf{d} \right )$.

In many parameter estimation problems in engineering, the forward model $f$ consists of solving a PDE. Take, as an example, estimating the heat conductivity of an iron rode by measuring the temperature at different parts of the rode at different times, for which the forward model is a transient heat conduction solver. In this case, computing the forward model $f$ in an MCMC simulation can be computationally expensive or even intractable, as usually MCMC samplers require thousands of millions of iterations to provide converged posterior distributions, and the model $f$ needs to be computed at each and every of these iterations. To alleviate this computational limitation, metamodels can serve as an approximation of the forward model. A variety of metamodels have been used in the literature to accelerate MCMC, including but not limited to, polynomial chaos expansions (e.g., \citep{marzouk2007stochastic,li2014adaptive,conrad2018parallel}), Gaussian process regression (e.g., \citep{overstall2013strategy,fielding2011efficient,rasmussen2003gaussian}), radial basis functions (e.g.,  \citep{bliznyuk2008bayesian}), data-driven deep neural networks (e.g., \citep{yan2019adaptive}), and physics-informed neural networks \citep{deveney2019deep}.

Physics-Informed Neural Networks (PINNs) are a class of deep neural networks that are trained, using automatic differentiation, to satisfy the governing laws of physics described in form of partial differential equations \citep{lagaris1998artificial,raissi2019physics}. The idea of using physics-informed training for  neural network solutions of differential equations  was first introduced in \cite{dissanayake1994neural,lagaris1998artificial,psichogios1992hybrid}, where neural networks solutions for initial/boundary value problems were developed. The method, however, did not gain much attention due to limitations in computational resources and optimization algorithms, until recently when researchers revisited this idea in (1) solving  challenging dynamic problems described by time-dependent nonlinear PDEs (e.g. \citep{raissi2017physics,raissi2019physics}), (2) solving variants of nonlinear PDEs  (e.g. \citep{berg2018unified,sirignano2017dgm,guo2019deep,weinan2018deep,goswami2019transfer,jagtap2019adaptive}), (3) data-driven discovery of PDEs (e.g. \citep{raissi2019physics,raissi2018deep,qin2018data,long2017pde}), (4) uncertainty quantification (e.g. \citep{raissi2019deep,raissi2018hiddena,zhu2019physics,meng2019composite,yang2019adversarial,kissas2019machine,xu2019neural}), (5) solving stochastic PDEs (e.g. \citep{yang2018physics,raissi2018forward,beck2018solving,weinan2017deep}), and (6) physics-driven regularization of DNNs (e.g. \cite{nabian2020physics}).

In \cite{nabian2019deep}, we introduced Physics-Informed Neural Networks for Uncertainty Quantification (PINN-UQ), which are the uncertainty-aware variant of the PINNs, and are used to effectively solve random PDEs. In this paper, we introduce a novel adaptive method (called APINN hereinafter) for efficient MCMC based on the PINN-UQ method. We specifically focus on a class of parameter estimation problems for which computing the likelihood function requires solving a PDE. The proposed method consists of: (1) constructing an offline PINN-UQ model as a global approximation to the forward model $f$; and (2) refining this global approximation model on the fly using samples from the MCMC sampler. We note that in \cite{deveney2019deep}, the authors have recently used offline PINNs as a surrogate to accelerate MCMC. However, it is commonly known that in Bayesian inference, the posterior distribution is usually concentrated on a small portion of the prior support \citep{li2014adaptive}. Therefore, constructing an accurate global approximate model for $f$ can be computationally challenging, especially for highly-nonlinear systems. On the other hand, as will be shown in the subsequent sections, the proposed APINN-MCMC method constantly refines this global approximation model on the fly, and also guarantees that the approximation error is always less than a user-defined residual error threshold.

The remainder of this paper is organized as follows. An introduction to the Metropolis-Hastings Algorithm, a popular variant of MCMC, is given in section \ref{C8_Bayesian}. Next, section \ref{sec:DNN} provides an introduction to the PINN-UQ method, which forms the foundation of the proposed APINN-MCMC method.  In section \ref{C8_APINN}, we describe the proposed APINN-MCMC method in detail. Next, section \ref{C8-Example} demonstrates the performance of the proposed method in solving a parameter estimation problem for a system governed by the Poisson equation. Finally, the last section concludes the paper.

\section{Metropolis-Hastings for Parameter Estimation} \label{C8_Bayesian}

In this study, without loss of generality, we focus on the Metropolis-Hastings (MH) algorithm \citep{metropolis1953equation,hastings1970monte}, which is the most popular variant of the MCMC methods \citep{andrieu2003introduction}. Metropolis algorithm is selected among the "10 algorithms with the greatest influence on the development and practice of science and engineering in the 20th century" by the IEEE's Computing in Science \& Engineering Magazine \citep{dongarra2000guest}. In Metropolis-Hastings algorithm, we construct a Markov chain for which, after a sufficiently large number of iterations, its stationary distribution converges almost surely to the posterior density $\pi\left ( \mathbf{p} | \mathbf{d} \right )$, and the states of the chain are then realizations from the parameters $\mathbf{p}$ according to the posterior distribution.

Let $q(\mathbf{z}^*| \mathbf{z}_{k-1})$ be a proposal distribution that generates a random candidate ${z}^*$ when the state of the chain is at the previous accepted sample ${z}_{k-1}$. In Metropolis-Hastings, we accept this sample with a probability defined as

\begin{equation}
\alpha = \min\left \{ 1, \frac{\pi(\textbf{z}^*)\pi \left ( \mathbf{d} | \mathbf{z}^* \right ) q(\mathbf{z}_{k-1}| \mathbf{z}^* )}{\pi(\textbf{z}_i)\pi \left ( \mathbf{d} | \mathbf{z}_i \right ) q(\mathbf{z}^*| \mathbf{z}_{k-1})} \right \}.
\end{equation}
Upon acceptance, the state of the chain transits to the accepted state ${z}^*$, or otherwise, remains unchanged. Theoretical convergence of the chain's stationary distribution to the posterior density $\pi\left ( \mathbf{p} | \mathbf{d} \right )$ is independent of the choice of proposal distribution $q$, and therefore, various options are possible. Among those, Gaussian or normal distribution seems to be the most commonly used proposal in the literature. The term $ \pi \left ( \mathbf{d} | \mathbf{z}^* \right )$ represents the likelihood of observations given the candidate state $z^*$, and usually consists of solving a forward model $f$. In this paper, we consider a class of problems for which this forward model is in the form of a PDE solver. Details of the Metropolis-Hastings sampler is summarized in Algorithm \ref{C8_MH_Algorithm}.

\begin{algorithm}
	\caption{Standard Metropolis-Hastings}\label{C8_MH_Algorithm}
	\begin{algorithmic}[1]
		\State Collect the measurements at $\mathbf{x}_m=\{\mathbf{x}_1, \cdots, \mathbf{x}_{n_m}\}$.
		\State Choose initial state $\textbf{z}_1$ and total number of samples  $N$.
		\State Choose a proposal distribution $q(\cdot)$.
		\For {$k=1:N -1$}
		\State Draw proposal $\textbf{z}^{*} \sim q(\cdot|\textbf{z}_{k})$.
		\State Compute the system response $u (\bm{x},\bm{{z}^{*}}) \,\,\, \forall \bm{x} \in \mathbf{x}_m$.
		\State Calculate the likelihood function $ \pi \left ( \mathbf{d} | \mathbf{z}^* \right )$.
		\State Calculate acceptance probability $\alpha = \min\left \{ 1, \frac{\pi(\textbf{z}^*)\pi \left ( \mathbf{d} | \mathbf{z}^* \right ) q(\mathbf{z}_{k}| \mathbf{z}^* )}{\pi(\textbf{z}_k)\pi \left ( \mathbf{d} | \mathbf{z}_k \right ) q(\mathbf{z}^*| \mathbf{z}_{k})} \right \}$.
		\State Draw $r_u \sim \text{Uniform} \left(0,1\right) $.
		\If {$r_u<\alpha$}
		\State $\textbf{z}_{k+1} = \textbf{z}^*$.
		\Else {}
		\State {$\textbf{z}_{k+1} = \textbf{z}_k$}.
		\EndIf
		\State $k = k+1$
		\EndFor
	\end{algorithmic}
\end{algorithm}

In Metropolis-hastings, we usually consider a user-defined burn-in period $t_b$, for which the first $t_b$ accepted samples are discarded in order to ensure that the remaining accepted samples are generated from the stationary distribution of the chain. Moreover, in order to prevent underflow, we usually compute the log-likelihood function instead of the likelihood function itself, and modify the steps in Algorithm \ref{C8_MH_Algorithm} accordingly.

\section{Theoretical Background on Physics-Informed Neural Networks for Uncertainty Quantification} \label{sec:DNN}

\subsection{ Feed-forward fully-connected deep neural networks}
Here,  a brief overview on feed-forward fully-connected deep neural networks is presented (a more detailed introduction can be found in \cite{lecun2015deep,goodfellow2016deep}). For notation brevity, let us first define the \textit{single hidden layer} neural network, since the generalization of the single hidden layer network to a network with multiple hidden layers, forming a \emph{deep} neural network, will be straightforward. Given the $d$-dimensional row vector $\bm{x} \in \mathbb{R}^{d}$ as model input, the $k$-dimensional output of a standard single hidden layer neural network is in the form of
\begin{equation} \label{Eqn.C2_OHL-NN}
\bm{y} = \sigma (\bm{x} \bm{W}_{1}+\bm{b}_{1}   ) \bm{W}_{2}+\bm{b}_{2},
\end{equation}
in which $\bm{W}_{1}$ and $\bm{W}_{2}$ are weight matrices of size $d\times q$ and $q\times k$, and $\bm{b}_{1}$ and $\bm{b}_{2}$ are \emph{bias} vectors of size $1\times q$ and $1\times k$, respectively. The function $\sigma( \cdot  )$ is an element-wise non-linear model, commonly known as the \textit{activation} function. In deep neural networks, the output of each activation function is transformed by a new weight matrix and a new bias, and is then fed to another activation function. With each new hidden layer in the neural network, a new set of weight matrices and biases is added to Equation (\ref{Eqn.C2_OHL-NN}). For instance, a feed-forward fully-connected deep neural network with three hidden layers is defined as
\begin{equation} \label{Eqn.C2_MHL-NN}
\bm{y} = \sigma \left(\sigma \left(\sigma \left(\bm{x} \bm{W}_{1}+\bm{b}_{1}   \right) \bm{W}_{2}+\bm{b}_{2}\right)\bm{W}_{3}+\bm{b}_{3} \right) \bm{W}_{4}+\bm{b}_{4} ,
\end{equation}
in which  $\{\bm{W}_i\}_{i=1}^{4},  \{\bm{b}_i\}_{i=1}^{4}$ are the weights and biases, respectively. Generally, the capability of neural networks to approximate complex nonlinear functions can be increased by adding more hidden layers and/or increasing the dimensionality of the hidden layers. 

Popular choices of activation functions include Sigmoid, hyperbolic tangent (Tanh), and Rectified Linear Unit (ReLU). The ReLU activation function, one of the most widely used functions, has the form of $f( \theta  )=\max( 0,\theta  )$. However, second and higher-order derivatives of ReLUs is 0 (except at $\theta=0$). This limits its applicability in our present work which deals with differential equations consisting potentially of second or higher-order derivatives. Alternatively, Tanh or Sigmoid activations may be used for second or higher-order PDEs. Sigmoid activations are non-symmetric and restrict each neuron's output to the interval $[0,1]$, and therefore, introduce a systematic bias to the output of neurons. Tanh activations, however, are antisymmetric and overcome the systematic bias issue caused by Sigmoid activations by permitting the output of each neuron to take a value in the interval [-1,1]. Also, there are empirical evidences that training of deep neural networks with antisymmetric activations is faster in terms of convergence, compared to training of these networks with non-symmetric activations \cite{lecun1991second,lecun2012efficient}.

In a regression problem given a number of training data points, we may use a Euclidean loss function in order to calibrate the weight matrices and biases, as follows
\begin{equation} \label{MSE Loss}
J( \bm{\Theta};\bm{X},\bm{Y})=\frac{1}{2M}\sum_{i=1}^{M}\left \| \bm{y}_i-\hat{\bm{y}}_{i} \right \|^{2},
\end{equation}
where $J$ is the mean square error divided by 2, $\bm{X}=\left \{ \bm{x}_1,\bm{x}_2,...,\bm{x}_M \right \}$ is the set of $M$ given inputs, $\bm{Y}=\left \{ \bm{y}_1,\bm{y}_2,...,\bm{y}_M \right \}$ is the set of $M$ given outputs, and $\left \{ \hat{\bm{y}}_1,\hat{\bm{y}}_2,...,\hat{\bm{y}}_M \right \}$ is the set of neural network predicted outputs calculated at the same set of given inputs $\bm{X}$. 

The model parameters can be calibrated according to
\begin{equation} \label{minimize_loss}
( \bm{W}_{1}^{*},\bm{W}_{2}^{*},\cdots,\bm{b}_{1}^{*},\bm{b}_{2}^{*},\cdots  )=\underset{{( \bm{W}_{1},\cdots,\bm{b}_{1}\cdots  )}}{\operatorname{argmin}} J(\bm{\Theta};\bm{X},\bm{Y}).
\end{equation}
This optimization is performed iteratively using Stochastic Gradient Descent (SGD) and its variants \cite{bottou2012stochastic,kingma2014adam,duchi2011adaptive,zeiler2012adadelta,sutskever2013importance}. Specifically, at the $i^{th}$ iteration, the model parameters $\bm{\Theta}=\{ \bm{W}_{1},\bm{W}_{2},\cdots,\bm{b}_{1},\bm{b}_{2},\cdots  \}$ are updated according to

\begin{equation} \label{descent step}
\bm{\Theta}^{(i+1)} = \bm{\Theta}^{(i)} - \eta^{(i)} \nabla_{\bm{\Theta}}J^{(i)}(\bm{\Theta}^{(i)}; \bm{X},\bm{Y}),
\end{equation}
where $\eta^{(i)}$ is the step size in the $i^{\textit{th}}$ iteration. The gradient of loss function with respect to model parameters $\nabla_{\bm{\Theta}} J$ is usually computed using \emph{backpropagation} \cite{lecun2015deep}, which is a special case of the more general technique called reverse-mode automatic differentiation \cite{baydin2018automatic}. In simplest terms, in backpropagation, the required gradient information is obtained by the backward propagation of the sensitivity of objective value at the output, utilizing the chain rule successively to compute partial derivatives of the objective with respect to each weight \cite{baydin2018automatic}. In other words, the gradient of  last layer is calculated first and the gradient of  first layer is calculated last. Partial gradient computations for one layer are reused in the gradient computations for the foregoing layers. This backward flow of information facilitates efficient computation of the gradient at each layer of the deep neural network \cite{lecun2015deep}. Detailed discussions about the backpropagation algorithm can be found in  \cite{goodfellow2016deep,lecun2015deep,baydin2018automatic}.

\subsection{Physics-informed neural networks for uncertainty quantification} \label{framework}

 In this paper, for brevity, we introduce the strong form of PINN-UQ, and the details for the variational form can be found in \cite{nabian2019deep}. We seek to calculate the approximate  solution $u(t,\bm{x},\bm{p}; \bm{\theta} )$  for the following differential equation

\begin{equation}\label{eqn:pde}
\begin{aligned}
\mathcal{L}( t,\bm{x}, \bm{p} ; u (t,\bm{x},\bm{p}; \bm{\theta} ) ) =0, \; \; \; \; & t \in [ 0,T ], \bm{x}\in \mathcal{D},  \bm{p}\in \mathbb{R}^d, \\
\mathcal{I}( \bm{x}, \bm{p} ; u (0,\bm{x},\bm{p}; \bm{\theta} ) )=0, \; \; \; \; & \bm{x}\in \mathcal{D}, \bm{p}\in \mathbb{R}^d,   \\
\mathcal{B}( t,\bm{x}, \bm{p} ; u (t,\bm{x},\bm{p}; \bm{\theta} ) )=0, \; \; \; \;  & t \in [ 0,T ], \bm{x}\in \mathcal{\partial {D}}, \bm{p}\in \mathbb{R}^d, 
\end{aligned}
\end{equation}
where $\theta$ include the parameters of the  function form of the solution,   $\mathcal{L}(\cdot)$ is a general differential operator that may consist of time derivatives, spatial derivatives, and linear and nonlinear terms, $\bm{x}$ is a position vector defined on a bounded continuous spatial domain $\mathcal{D} \subseteq \mathbb{R}^D , D \in \left \{ 1,2,3 \right \} $ with boundary $\mathcal{\partial {D}}$, $t \in \left[ 0,T \right]$, and $\bm{p}$ denotes an $\mathbb{R}^d$-valued random vector, with a joint distribution $\rho_{\bm{p}}$, that characterizes input uncertainties. Also, $\mathcal{I}(\cdot)$ and $\mathcal{B}(\cdot)$ denote, respectively, the initial and boundary conditions and may consist of differential, linear, or nonlinear operators. In order to calculate the solution, i.e. calculate the parameters  $\bm{\theta}$, let us consider the   following non-negative residuals, defined over the entire spatial, temporal and stochastic domains 
\begin{equation}\label{eqn:l2-redidual}
\begin{aligned}
r_\mathcal{L} (  \bm{\theta}  ) &=\int_{\left[ 0,T \right] \times \mathcal{D} \times \mathbb{R}^d  }( \mathcal{L} (  t,\bm{x}, \bm{p}; \bm{\theta} ) )^2 \rho_{\bm{p}}  \, \diff t \, \diff \bm{x} \, \diff \bm{p},   \\
r_\mathcal{I} (  \bm{\theta}  ) &=\int_{\mathcal{D}\times \mathbb{R}^d  }( \mathcal{I} (  \bm{x}, \bm{p}; \bm{\theta} ) )^2 \rho_{\bm{p}}  \,\diff \bm{x} \, \diff \bm{p},   \\
r_\mathcal{B} (  \bm{\theta}  )&=\int_{\left[ 0,T \right] \times \mathcal{\partial {D}}\times \mathbb{R}^d }( \mathcal{B} (  t,\bm{x}, \bm{p}; \bm{\theta} ) )^2 \rho_{\bm{p}}  \, \diff t \, \diff \bm{x} \, \diff \bm{p}.   
\end{aligned}
\end{equation}
The optimal  parameters $\bm{\theta^*}$ can then be calculated according to 
\begin{equation} \label{eqn:theta-star}
\begin{aligned}
\bm{\theta^*}=\underset{{ \bm{\theta} }}{\operatorname{argmin}}\, r_\mathcal{L}( \bm{\theta}  ),   \\
\text{s.t.} \quad r_\mathcal{I} (  \bm{\theta}  )=0, \, r_\mathcal{B} (  \bm{\theta}  )=0.
\end{aligned}
\end{equation}
Therefore, the solution to the random differential equation defined in Equation \ref{eqn:pde} is reduced to an optimization problem, where initial and boundary conditions can   be viewed as constraints.  In PINN-UQ, the constrained optimization~\ref{eqn:theta-star} is reformulated as an unconstrained optimization with a modified loss function that also accommodate the constraints. To do so,  two different approaches are adopted, namely soft  and  hard assignment of constraints, which differ in  how strict the constraints are imposed \cite{marquez2017imposing}. In the soft assignment,  constraints are translated into additive penalty terms in the loss function (see e.g. \cite{sirignano2017dgm}). This approach is easy to implement but it is not clear how to  tune the relative importance of different terms in the loss function, and also there is no guarantee that the final solution will satisfy the constraints. In the hard assignment of constraints, the function form of the approximate solution  is formulated   in such a way that any solution with that function form is guaranteed to  satisfy the conditions (see e.g. \cite{lagaris1998artificial}). Methods with hard  assignment of constraints are more robust compared to their soft counterparts. However, the constraint-aware formulation of the function form of the solution is not straightforward for boundaries with irregularities or for mixed boundary conditions (e.g. mixed Neumann and Dirichlet boundary conditions). In what follows, we explain how the approximate solution in the form of a DNN can be calculated using these two assignment approaches. Let us denote the solution obtained by a feed-forward fully-connected deep residual network  by $u_{\text{DNN}}(t,\bm{x},\bm{p}; \bm{\theta} )$. The inputs to this deep residual network are $t$, $\bm{x}$, and realizations from the random vector $\bm p$.

For soft assignment of constraints, we use a generic DNN form for the solution. That is, we set $u_s(t,\bm{x},\bm{p}; \bm{\theta} ):=u_{\text{DNN}}(t,\bm{x},\bm{p}; \bm{\theta} )$, and solve the following unconstrained optimization problem
\begin{equation} \label{loss-soft}
\bm{\theta^*}=\underset{{ \bm{\theta} }}{\operatorname{argmin}}\, \underbrace{r_\mathcal{L}(  \bm{\theta} ) +\lambda_1 r_\mathcal{I}  (  \bm{\theta}   )+\lambda_2 r_\mathcal{B}  (  \bm{\theta}   )}_{J_s(  \bm{\theta} ; u_s)},
\end{equation}
in which $\lambda_1$ and $\lambda_2$ are weight parameters, analogous to collocation finite element method in which weights are used to adjust the relative importance of each residual term \cite{bochev2006least}. 

In  hard assignment of constraints, the uncertainty-aware function form for the approximate solution can take the following general form \cite{lagaris1998artificial}.

\begin{equation} \label{trial-function}
u_h (t,\bm{x},\bm{p}; \bm{\theta} )=C( t, \bm{x})+G( t,\bm{x},u_{\text{DNN}}(t,\bm{x},\bm{p}; \bm{\theta} )),
\end{equation}
where $C ( t, \bm{x} )$ is a function that satisfies the initial and boundary conditions and has no tunable parameters, and, by construction, $G( t,\bm{x},u_{\text{DNN}} (t,\bm{x},\bm{p}; \bm{\theta}  ) )$ is derived such that it has no contribution to the initial and boundary conditions. A systematic way to construct the functions $C(  \cdot )$ and $G(  \cdot)$ is presented in \cite{lagaris1998artificial}.  Our goal is then to estimate the DNN parameters $\bm \theta$ according to 
\begin{equation} \label{loss-hard}
\bm{\theta^*}=\underset{{ \bm{\theta} }}{\operatorname{argmin}}\, \underbrace{r_\mathcal{L} (  \bm{\theta}  )}_{J_h(  \bm{\theta};u_h )}.
\end{equation}

To solve the two unconstrained optimization problems ~\ref{loss-soft} and~\ref{loss-hard}, we make use of stochastic gradient descent (SGD) optimization algorithms \cite{ruder2016overview}, which are a variation of gradient-descent algorithms. In each iteration of an SGD algorithm,  the gradient of loss function is approximated using only one point in the input space, based on which the neural network parameters are updated. This iterative update is shown to result in an unbiased estimation of the gradient, with bounded variance \cite{bottou2010large}.

Specifically, in soft assignment of constraints, on the $i^{\textit{th}}$ iteration, the DNN parameters are updated according to
\begin{equation} \label{ldescent step-soft}
\bm{\theta}^{(i+1)} = \bm{\theta}^{(i)} - \eta^{(i)} \nabla_{\bm{\theta}}\tilde{J}_s^{(i)}(\bm{\theta}),
\end{equation}
where $\eta^{(i)}$ is  the step size in the $i^{\textit{th}}$ iteration, and $\tilde{J}_s^{(i)}(\bm{\theta})$ is the  approximate loss function, obtained by numerically evaluating integrals in Equations~\ref{eqn:l2-redidual} using a single   sample point. That is, 
\begin{multline} \label{loss-approximate-soft}
\tilde{J}_s^{(i)}(\bm{\theta})= \left[\mathcal{L}( t^{(i)},\bm{x}^{(i)},\bm{p}^{(i)}; u_s(t^{(i)},\bm{x}^{(i)},\bm{p}^{(i)} ;\bm{\theta} ) ) \right]^2 + \\
\lambda_1 \left[\mathcal{I}( \bm{x}^{(i)},\bm{p}^{(i)}; u_s(0,\bm{x}^{(i)},\bm{p}^{(i)} ;\bm{\theta} ) ) \right]^2 
+\lambda_2 \left[\mathcal{B}( t^{(i)},\bm{\underbar{x}}^{(i)},\bm{p}^{(i)}; u_s(t^{(i)},\underbar{$\bm{x}$}^{(i)},\bm{p}^{(i)} ;\bm{\theta} ) ) \right]^2.
\end{multline}
where $t^{(i)},\bm{x}^{(i)}$ and $\underbar{$\bm{x}$}^{(i)}$ are  uniformly drawn in $\left[ 0,T \right], \mathcal{D}$ and $\mathcal{\partial{D}}$, and $\bm{p}^{(i)}$ is drawn in  $\mathbb{R}^d $ according to distribution $\rho_{\bm{p}}$. The gradient of loss function with respect to model parameters $\nabla_{\bm{\Theta}} \tilde{J}_s$ can be calculated using backpropagation \cite{baydin2018automatic}. The term $\mathcal{L}( t^{(i)},\bm{x}^{(i)},\bm{p}; u_s(t^{(i)},\bm{x}^{(i)},\bm{p}^{(i)} ;\bm{\theta})) $ also involves gradients of the solution $u_s$ with respect to $t$ and $\bm{x}$, which may be calculated using reverse-mode automatic differentiation.

Similarly, in hard assignment of constraints, the DNN parameters are updated according to 
\begin{equation} \label{ldescent step-hard}
\bm{\theta}^{(i+1)} = \bm{\theta}^{(i)} - \eta^{(i)} \nabla_{\bm{\theta}}\tilde{J}_h^{(i)}(\bm{\theta}),
\end{equation}
where
\begin{equation} \label{loss-approximate-hard}
\tilde{J}_h^{(i)}(\bm{\theta}  )=\left[\mathcal{L}( t^{(i)},\bm{x}^{(i)},\bm{p}^{(i)}; u_h(t^{(i)},\bm{x}^{(i)},\bm{p}^{(i)} ;\bm{\theta}  ) ) \right]^2.
\end{equation}

It is common in practice that in each iteration the gradient of the loss function is calculated at and averaged over $n$ different sample input points instead of being evaluated at only one point. Such approaches are called mini-batch gradient descent algorithms \cite{ruder2016overview}, and compared to stochastic gradient descent algorithms, are more robust and more   efficient. 

Algorithm \ref{Algorithm1} summarizes the proposed step-by-step approach. 
The algorithm can be stopped based on a pre-specified  maximum number of iterations (as shown in Algorithm~\ref{Algorithm1}, or using an on-the-fly stoppage criteria based on variations in  the loss function values across a few iterations. For brevity, $\tilde{J}( \bm{\theta} )$ represents the loss function regardless of hard or soft assignment of constraints.

\begin{algorithm}
	\caption{Physics-Informed Neural Networks for Uncertainty Quantification}\label{Algorithm1}
	\begin{algorithmic}[1]
\State Set the DNN architecture (number of layers, dimensionality of each layer, and activation function; and for residual networks, also the structure of shortcut connections).
\State Initialize DNN parameters $\bm{\theta}^{(0)}$.
\State Select a method for assignment of constraints.
\State Form the target function $u (\bm{x},\bm{p}; \bm{\theta} )$ according to Equation \ref{trial-function} or  \ref{loss-approximate-soft}.
\State Form the loss function $\tilde{J}( \bm{\theta} )$ according to Equation \ref{loss-approximate-hard} or  \ref{loss-approximate-soft}.
\State For the mini-batch gradient descent algorithm, specify optimizer hyper-parameters and batch size $n$.
\State Specify  maximum number of iterations $i_{\text{max}}$; set $i=0$.
\While {${i<i_{\text{max}}}$ }
\State Generate $n$ random inputs $\{\bm{x}_j^{(i)},\bm{p}_j^{(i)} \}_{j=1}^{n}$, sampled uniformly from $\mathcal{D}$, and from $\mathbb{R}^{n_p} $ according to
\Statex[1] $\pi\left ( \mathbf{p} \right )$ (and $\{\bm{\bar{x}}_j^{(i)}\}_{j=1}^{n}$ uniformly from $\mathcal{\partial{D}}$, in soft assignment of constraints).
\State Take a descent step  $\bm{\theta}^{(i+1)} = \bm{\theta}^{(i)} - \eta^{(i)} \nabla_{\bm{\theta}}\tilde{J}^{(i)}$; $i = i+1$
\EndWhile
		
	\end{algorithmic}
\end{algorithm}

\section{Adaptive Physics-Informed Neural Networks (APINNs) for Markov Chain Monte Carlo} \label{C8_APINN}

In this study, we propose to use PINN-UQ as an approximation to the forward model $f$, which consists of solving a PDE (or a system of PDEs) characterized by uncertain parameters. In approximating the forward model $f$, we ideally want to have control over the approximation error to make sure the ultimate posterior density results are reliable. Therefore, we need to train a PINN-UQ as a representative of the forward model $f$ such that, for each point in the coupled spatial, temporal, and stochastic spaces,  the residual error is less than a user-defined threshold $\epsilon_t$.  However, in Bayesian inference, it is commonly known that the posterior density can reside on a small fraction of the prior support, and therefore, training a sufficiently accurate PINN-UQ over the entire prior support can be inefficient and challenging, and more importantly unnecessary, as implemented in \cite{deveney2019deep}. In this work, we introduce the Adaptive Physics-Informed Neural Networks (APINNs), which are PINN-UQ models that are adaptively refined to meet an error threshold as required. Instead of training a sufficiently-accurate PINN-UQ over the entire prior support such that the residual error is less than the required threshold $\epsilon_t$, in APINNs, we relax this requirement, and train a PINN-UQ with only a viable accuracy. Next, we run our MH sampler, and for each parameter candidate $z^*$, we take a few training iterations in order to reduce the residual error to meet the threshold $\epsilon_t$, only for that candidate realization. 

In adaptively refining the global PINN-UQ as MCMC runs, there are two extremes that can be considered for updating the PINN-UQ model parameters. One extreme is to discard all changes to the model parameters of the PINN-UQ; the changes are only made to ensure the residual error is less than $\epsilon_t$ at $z^*$, and are then discarded, meaning that the model parameters are restored to that of the offline PINN-UQ. The second extreme is to constantly update and keep the changes to the model parameters as the MCMC sampler proceeds. There are downsides to both of these approaches. The first approach is inefficient, as none of the computational effort in online training is reflected in the global model. In the second approach, excessive local refinement of the APINN can adversely affect the global accuracy of the APINN. In this study, we take a different approach. At each candidate $z^*$, we refine the model parameters as needed, but only keep the parameter update for the first iteration. This means that we start the training of our global approximating model using samples draw from the prior distribution of parameters (offline phase), but later we refine this global approximation model using samples drawn from the posterior distribution (online phase).

\begin{figure}[h]
	\begin{center}
		\includegraphics[width=0.7\linewidth]{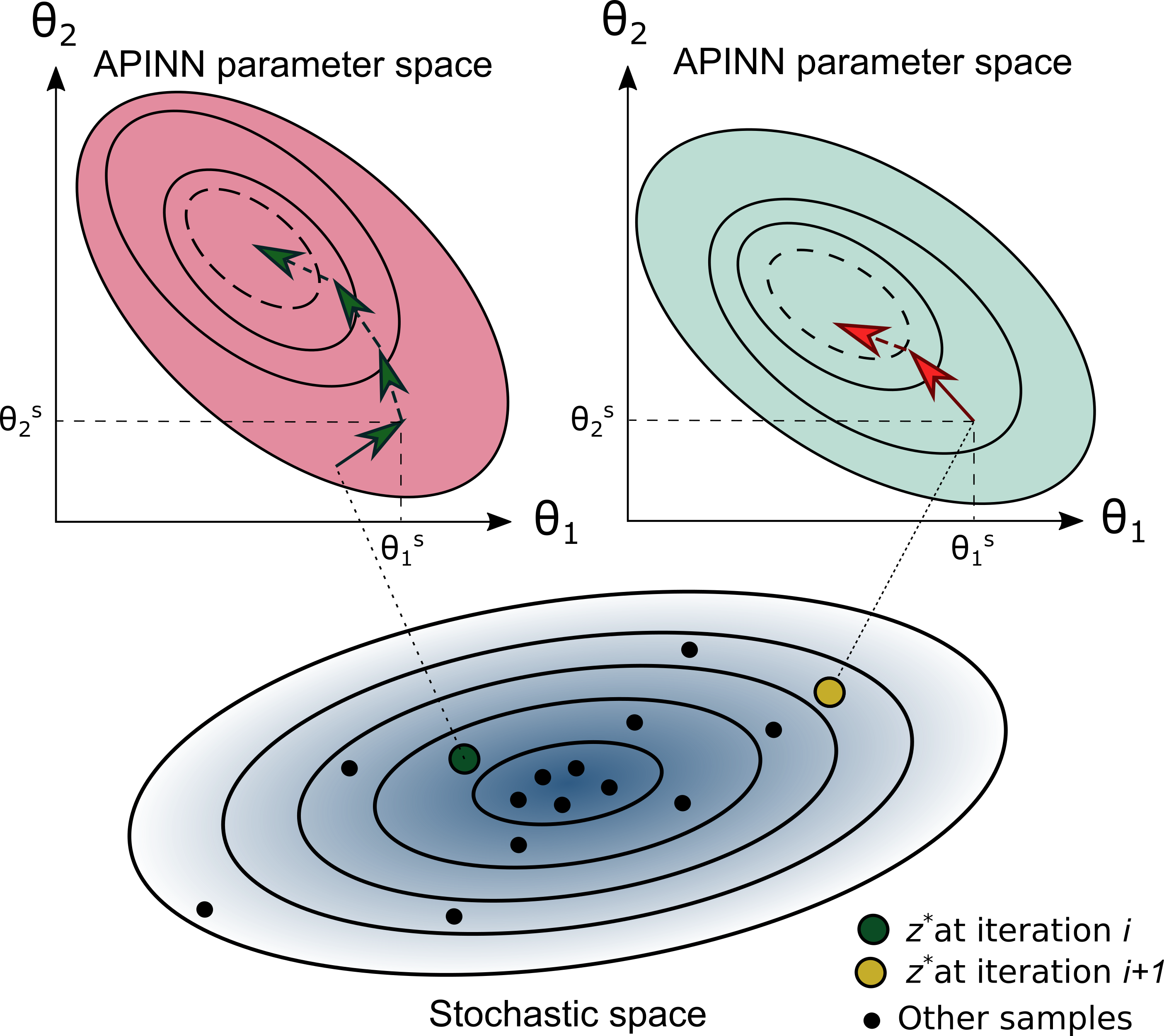}
		\caption{A schematic of the parameter update rule in the APINNs for MCMC. The stochastic space is depicted on the bottom, with the contour map showing the posterior distribution. On the top, the APINN model parameter space is depicted for two consecutive and different realizations of the stochastic space, with the contour map showing the expected value of the APINN loss over the physical domain. For the candidate $z^*$ at iteration $i$, four training steps are taken to ensure that the average residual error of APINN (over the physical domain for the specific value of $z^*$ at iteration $i$) is less than the threshold $\epsilon_t$. After the refined APINN solution is computed, as shown by the solid arrow, only the first training step is saved (that is, the APINN model parameters set to $\theta_1^s, \theta_2^s$), and the rest are discarded (dashed arrows). For the new candidate $z^*$ at iteration $i+1$, two training steps are taken to refine the APINN model parameters and satisfy the residual error threshold $\epsilon_t$, starting from the parameters values after the first training iteration of the previous round of refinements (i.e., $\theta_1^s, \theta_2^s$). Again, after the refined APINN solution is computed, only the first training step is saved (solid arrow), and the second one is discarded (dashed arrow).}
		\label{C8_fig_sketch}
	\end{center}
\end{figure}

Figure \ref{C8_fig_sketch} represents a schematic of the parameter update rule in the APINNs for MCMC. The stochastic space is depicted on the bottom, with the contour map showing the posterior distribution. On the top, the APINN model parameter space is depicted for two consecutive and different realizations of the stochastic space, with the contour map showing the expected value of the local approximation loss over the physical domain. For the candidate $z^*$ at iteration $i$, four training steps are taken to ensure that the average residual error of the local approximating model (over the physical domain for the specific value of $z^*$ at iteration $i$) is less than the threshold $\epsilon_t$. After the refined solution is computed, as shown by the solid arrow, only the first training step is saved (that is, the APINN model parameters are set to $\theta_1^s, \theta_2^s$), and the rest are discarded (dashed arrows). For the new candidate $z^*$ at iteration $i+1$, two training steps are taken to refine the APINN model parameters and satisfy the residual error threshold $\epsilon_t$, starting from the parameters values after the first training iteration of the previous round of refinements (i.e., $\theta_1^s, \theta_2^s$). Again, after the refined solution is computed, only the first training step is saved (solid arrow), and the second one is discarded (dashed arrow).

Algorithm \ref{C8_APINN_Algorithm} summarizes the steps for the proposed APINN method, based on the MH variant of MCMC. For brevity, $\tilde{J}( \bm{\theta} )$ represents the loss function regardless of hard or soft assignment of constraints. The algorithm consists of two parts of offline and online training. The line numbers that are represented in boldface denote the steps that are recommended to be executed on GPU for computational efficiency. 

\begin{algorithm}[h]
	\caption{APINNs for Metropolis-Hastings}\label{C8_APINN_Algorithm}
	\begin{algorithmic}[1]
		\State Generate an offline PINN-UQ approximate model for $f$ using Algorithm \ref{Algorithm}.
		\State Collect the measurements at $\mathbf{x}_m=\{\mathbf{x}_1, \cdots, \mathbf{x}_{n_m}\}$.
		\State Choose initial state $\textbf{z}_1$ and total number of samples $N$, and the surrogate error tolerance $\epsilon_t$.
		\State Choose a proposal distribution $q(\cdot)$.
		\For {$k=1:N -1$}
		\State Draw proposal $\textbf{z}^{*} \sim q(\cdot|\textbf{z}_{k})$.
		\boldnext
		\State Compute the system response $\tilde{u} (\bm{x},\bm{{z}^{*}}; \bm{\theta}^{(i)} ) \,\,\, \forall \bm{x} \in \mathbf{x}_m$.
		\boldnext
		\If {any \{$\tilde{u} (\bm{x},\bm{{z}^{*}})\}_{\forall \bm{x} \in \mathbf{x}_m}$ > $\epsilon_t$}
		\boldnext
		\State Generate $n$ random inputs $\{\bm{x}_j^{(i)},\bm{p}_j^{(i)} \}_{j=1}^{n}$, sampled uniformly from $\mathcal{D}$, and 
		\Statex[2] from $\mathbb{R}^{n_p} $ according to $\pi\left ( \mathbf{p} \right )$ (and $\{\bm{\bar{x}}_j^{(i)}\}_{j=1}^{n}$ uniformly from $\mathcal{\partial{D}}$, in soft 
		\Statex[2] assignment of constraints).
		\boldnext
		\State Take a descent step  $\bm{\theta}^{(i+1)} = \bm{\theta}^{(i)} - \eta^{(i)} \nabla_{\bm{\theta}}\tilde{J}^{(i)}$; $i = i+1$; $c = 0$.
		\boldnext
		\State Compute the system response $\tilde{u} (\bm{x},\bm{{z}^{*}}; \bm{\theta}^{(i)} ) \,\,\, \forall \bm{x} \in \mathbf{x}_m$.
		\boldnext
		\While {any \{$\tilde{u} (\bm{x},\bm{{z}^{*}}; \bm{\theta}^{(i+c)})\}_{\forall \bm{x} \in \mathbf{x}_m}$ > $\epsilon_t$}
		\boldnext
		\State Generate $n$ random inputs $\{\bm{x}_j^{(i)},\bm{p}_j^{(i)} \}_{j=1}^{n}$, sampled uniformly from $\mathcal{D}$,
		\Statex and from $\mathbb{R}^{n_p} $ according to $\pi\left ( \mathbf{p} \right )$ (and $\{\bm{\bar{x}}_j^{(i)}\}_{j=1}^{n}$ uniformly from $\mathcal{\partial{D}}$, in 
		\Statex soft assignment of constraints).
		\boldnext
		\State Take a descent step  $\bm{\theta}^{(i+c+1)} = \bm{\theta}^{(i+c)} - \eta^{(i+c)} \nabla_{\bm{\theta}}\tilde{J}^{(i+c)}$; $c = c+1$.
		\boldnext
		\State Compute the system response $\tilde{u} (\bm{x},\bm{{z}^{*}}; \bm{\theta}^{(i+c)} ) \,\,\, \forall \bm{x} \in \mathbf{x}_m$.
		\EndWhile
		\EndIf
		\State Calculate the likelihood function $ \pi \left ( \mathbf{d} | \mathbf{z}^* \right )$.
		\State Calculate acceptance probability $\alpha = \min\left \{ 1, \frac{\pi(\textbf{z}^*)\pi \left ( \mathbf{d} | \mathbf{z}^* \right ) q(\mathbf{z}_{k}| \mathbf{z}^* )}{\pi(\textbf{z}_k)\pi \left ( \mathbf{d} | \mathbf{z}_k \right ) q(\mathbf{z}^*| \mathbf{z}_{k})} \right \}$.
		\State Draw $r_u \sim \text{Uniform} \left(0,1\right) $.
		\If {$r_u<\alpha$}
		\State $\textbf{z}_{k+1} = \textbf{z}^*$.
		\Else {}
		\State {$\textbf{z}_{k+1} = \textbf{z}_k$}.
		\EndIf
		\EndFor
		
	\end{algorithmic}
\end{algorithm}

\section{Numerical Example}\label{C8-Example}

In this section, we numerically demonstrate the performance of the proposed APINN method in solving a parameter estimation problem for a system governed by the Poisson equation. Throughout this example, DNN training is performed using TensorFlow \citep{abadi2016tensorflow} on a NVIDIA Tesla P100-PCIE-16GB GPU. The Adam optimization algorithm \citep{kingma2014adam} is used to find the optimal neural network parameters, with the learning rate, $\beta_1$ and $\beta_2$ set to 0.0001, 0.9, and 0.999, respectively.

\begin{figure}[h]
	\begin{center}
		\begin{subfigure}{0.44 \linewidth}
			\includegraphics[width=.99\linewidth]{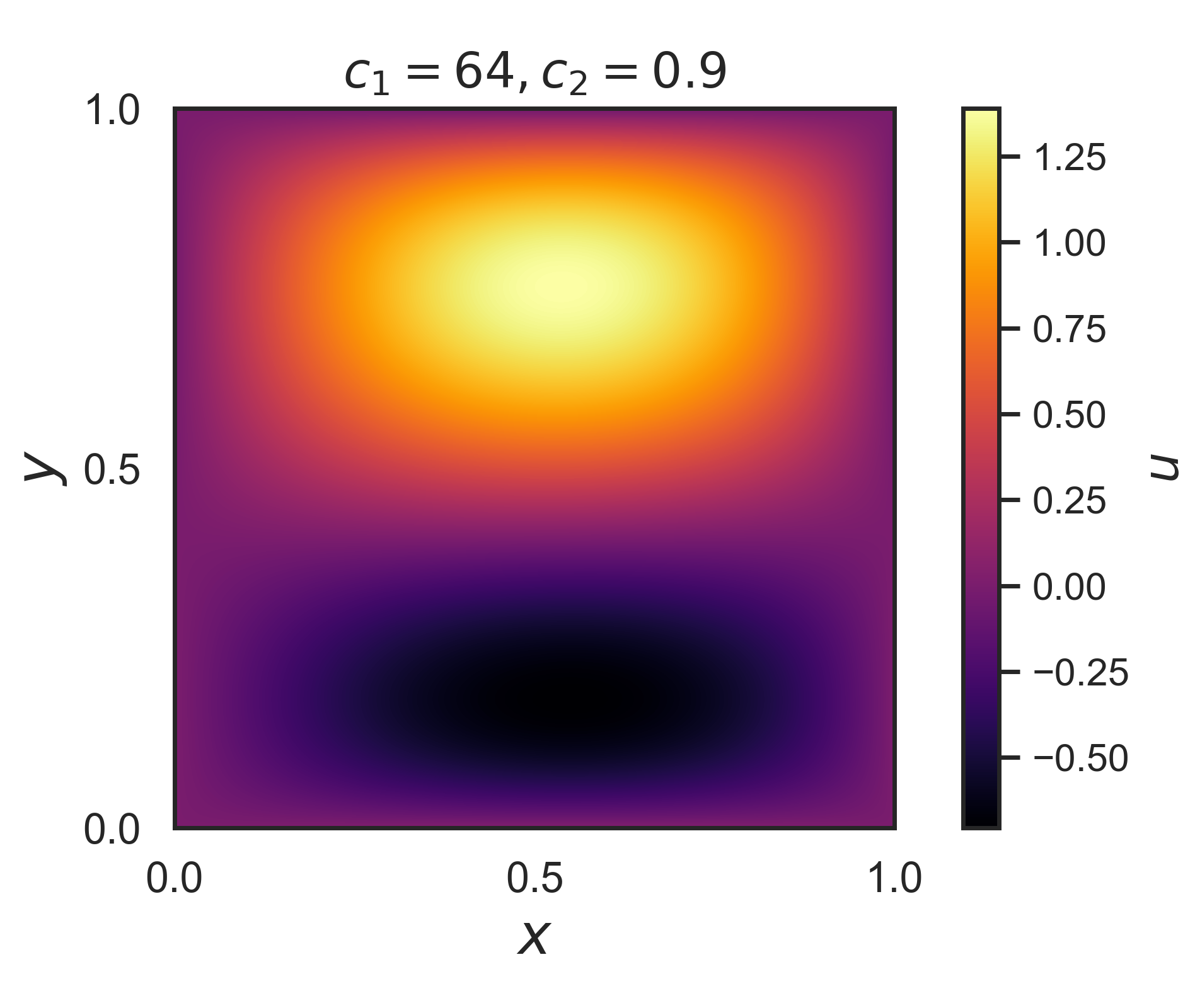}
			\centering	
		\end{subfigure}
		\quad
		\begin{subfigure}{0.44 \linewidth}
			\includegraphics[width=.99\linewidth]{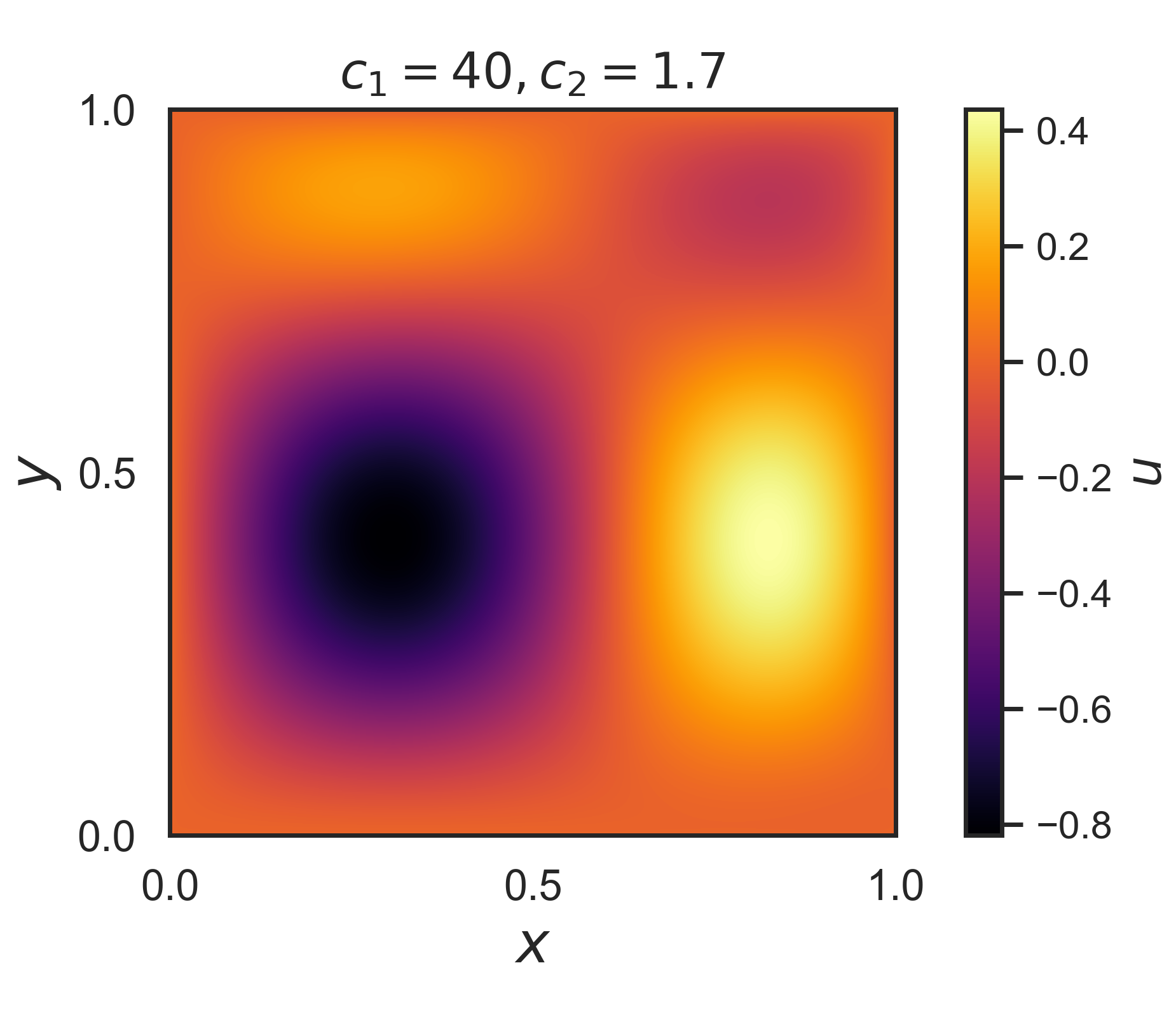}
		\end{subfigure}
		\\
		\begin{subfigure}{0.44 \linewidth}
			\includegraphics[width=.99\linewidth]{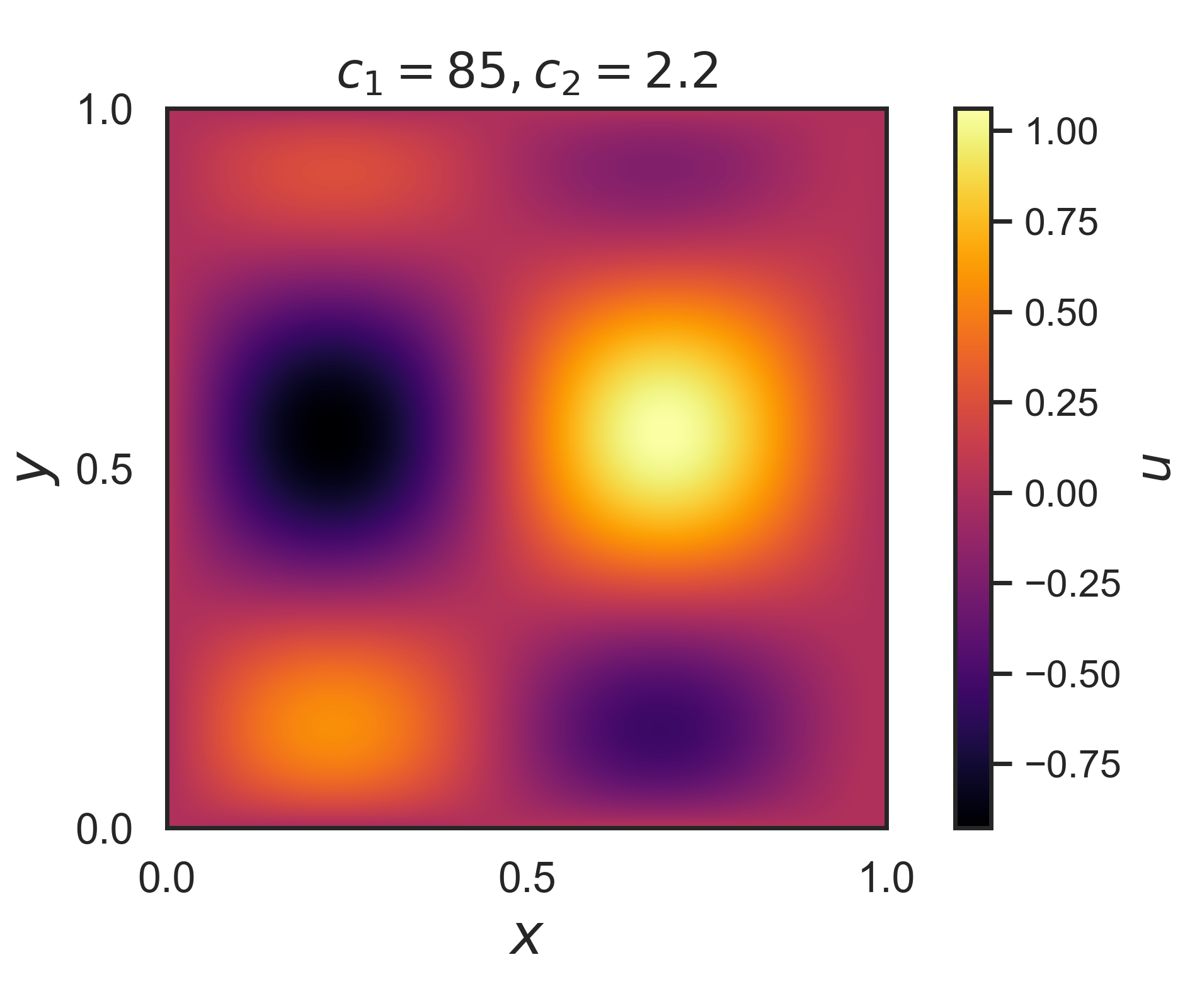}
		\end{subfigure}
		\quad
		\begin{subfigure}{0.44 \linewidth}
			\includegraphics[width=.99\linewidth]{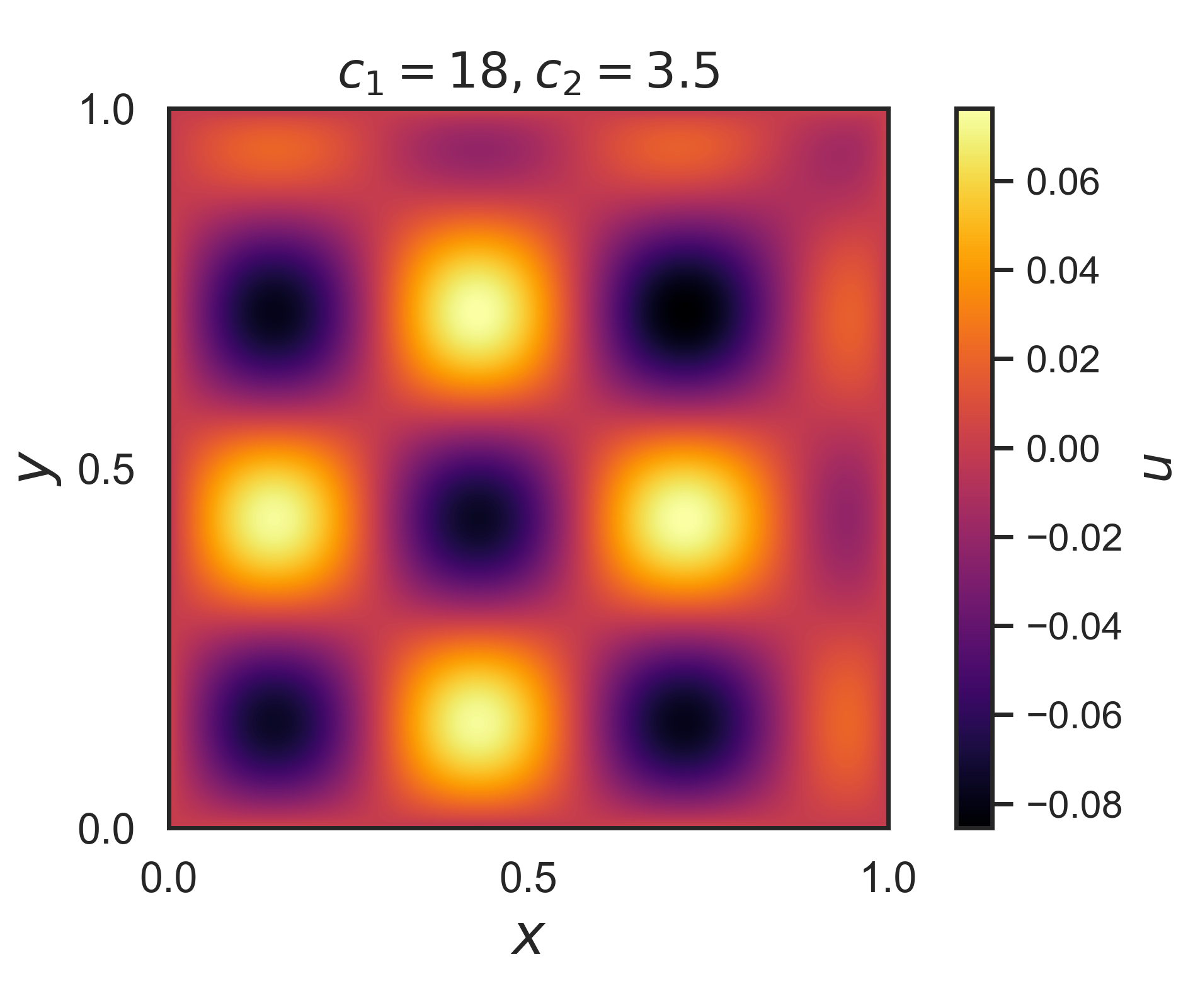}
		\end{subfigure}
		\captionsetup{}
		\caption{Four different realizations of the system response defined in Equation \ref{C8_Eq_PDE}.} 
		\label{C8_fig_realizations}
	\end{center}
\end{figure}

Let us consider a system governed by the following Poisson equation

\begin{align}\label{C8_Eq_PDE}
-\left( \frac{\partial ^2 u}{\partial x^2 } + \frac{\partial ^2 u}{\partial y^2 } \right) &= c_1 sin(c_2\pi x) cos(c_2\pi y), \nonumber\\
u( 0,y,c_1,c_2 )=0, \, u( x,0,c_1,c_2 )=0 &, u( 1,y,c_1,c_2 )=0, \, u( x,1,c_1,c_2 )=0,
\end{align}
where $u$ denotes the system response, $x \in\left [ 0,1 \right ]$, $y \in\left [ 0,1 \right ]$ are the spatial coordinates, and $c_1 \in \left [ 10,100 \right ]$, $c_2\in \left [ 0.1,4 \right] $ are the system parameters to be estimated. Figure \ref{C8_fig_realizations} shows four realizations of the system response $u$ for different choices of $c_1$ and $c_2$ values.

\begin{figure}
	\begin{center}
		\includegraphics[width=0.55\linewidth]{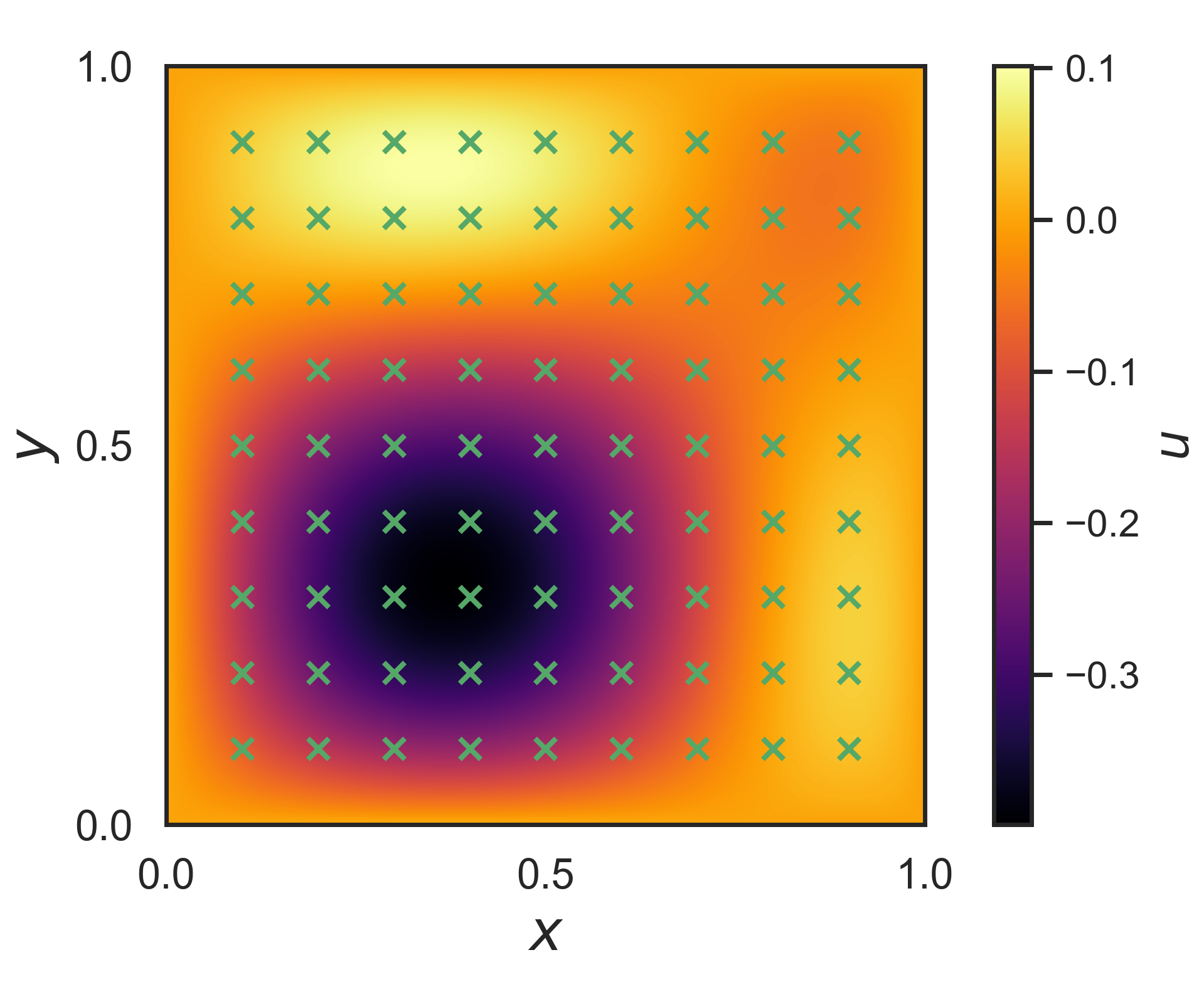}
		\caption{System response (Equation \ref{C8_Eq_PDE}) with $c_1$ and $c_2$ set to $15.0$ and $1.4$, respectively, and the location of measurement sensors.}
		\label{C8_fig_sensors}
	\end{center}
\end{figure}

We generate synthetic measurements as follows: (1) We set $c_1$ and $c_2$, respectively,  to $15.0$ and $1.4$, and, using the Finite Difference method, we compute the system response at 81 sensor locations uniformly distributed across the spatial domain, as shown in Figure \ref{C8_fig_sensors}; and (2) we add a zero-mean normally-distributed noise, with a standard deviation of 6\% of the 2-norm of system response at sensor locations. Note that, from this point forward, we assume we only have access to the noisy measurements at sensor locations, and the true value of the parameters $c_1$ and $c_2$ is assumed unknown.

To solve this parameter estimation problem, we run three separate MH samplers: (1) MH-FD, for which the likelihood function is computed using the Finite Difference method; (2) MH-PINN-UQ, for which the likelihood function is computed using an offline PINN-UQ; and (3) MH-APINN, for which the likelihood function is computed using the proposed APINN. We assume a uniform prior for parameters $c_1$ and $c_2$. A normal distribution is selected as the proposal distribution, with a covariance of $\left([  3.2, 0], [0, 0.006]\right)$. The initial state is set to 45 and 1.95 for $c_1$ and $c_2$, respectively, and the samplers are run for 50,000 iterations, with burn-in period set to 1,000.

\begin{figure}[h]
	\begin{center}
		\begin{subfigure}{0.48 \linewidth}
			\includegraphics[width=.99\linewidth]{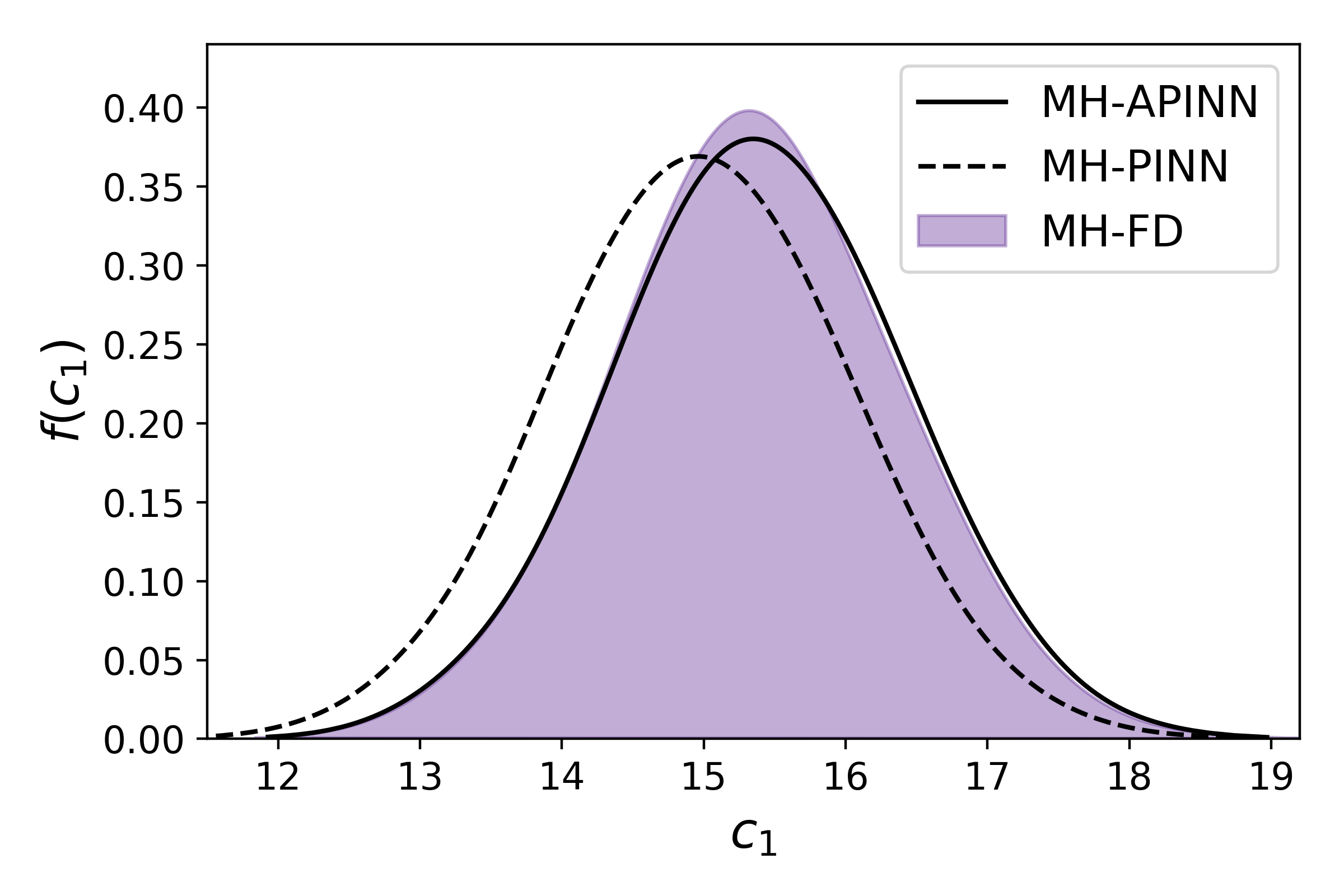}
			\caption{}
			\centering	
		\end{subfigure}
		\quad
		\begin{subfigure}{0.48 \linewidth}
			\includegraphics[width=.99\linewidth]{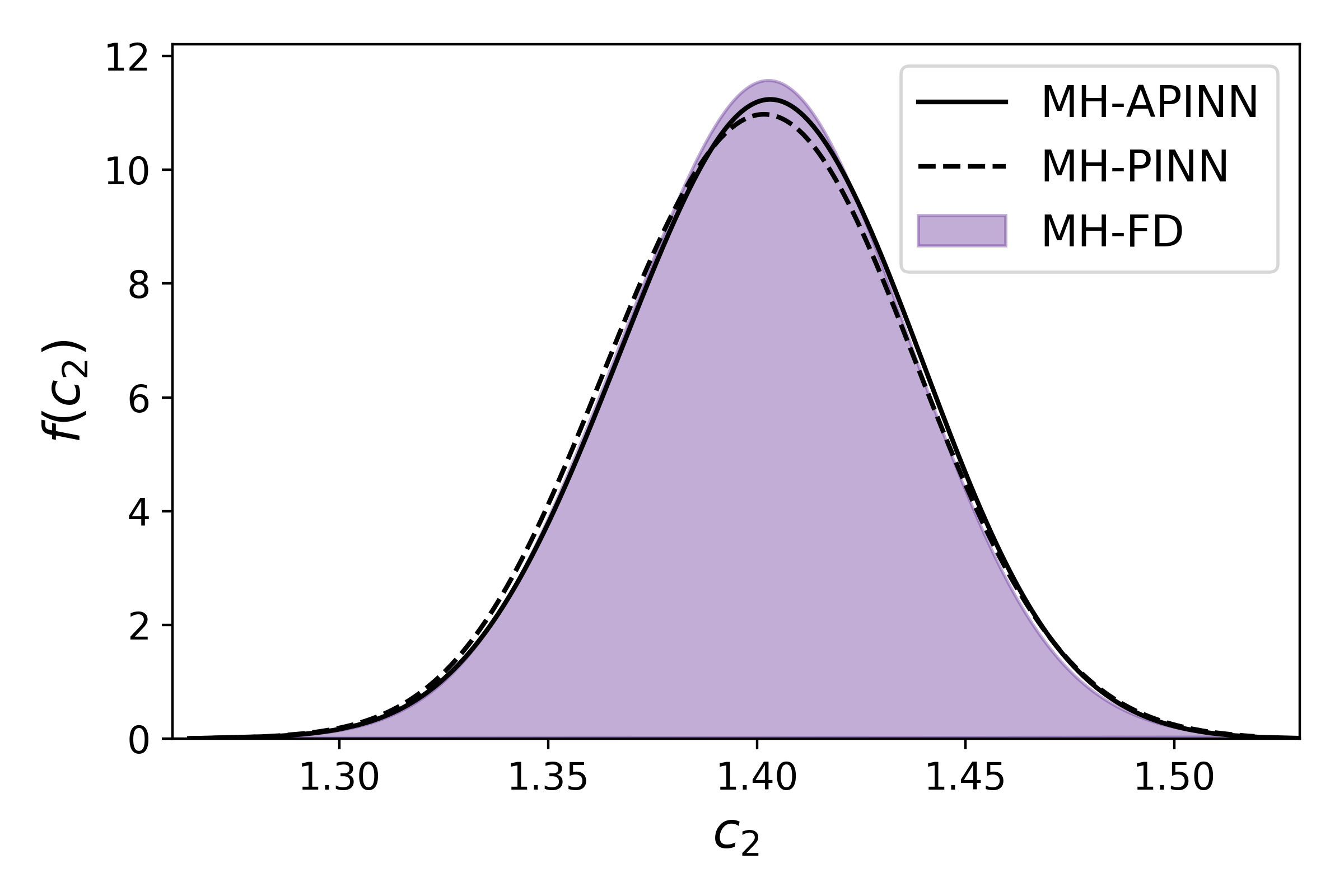}
			\caption{}
		\end{subfigure}
		\captionsetup{}
		\caption{Estimated posterior distributions: (a) marginal distribution for $c_1$; (b) marginal distribution for $c_2$. Results for MH-FD, MH-PINN, and MH-APINN methods are shown in shades, dashed line, and solid line, respectively.} 
		\label{C8_fig_density_marginal}
	\end{center}
\end{figure}

\begin{figure}[h]
	\begin{center}
		\begin{subfigure}{0.48 \linewidth}
			\includegraphics[width=.99\linewidth]{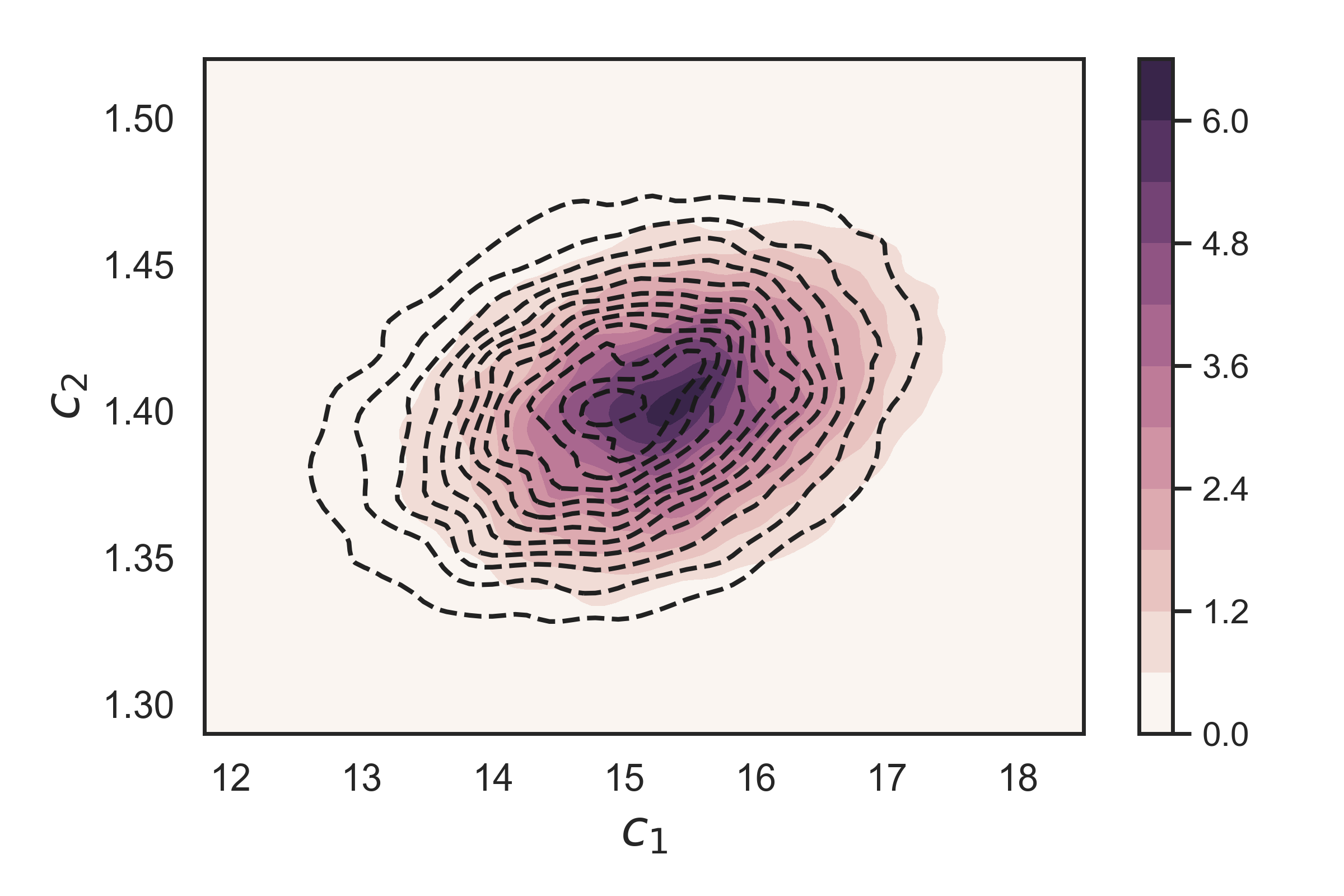}
			\caption{}
			\centering	
		\end{subfigure}
		\quad
		\begin{subfigure}{0.48 \linewidth}
			\includegraphics[width=.99\linewidth]{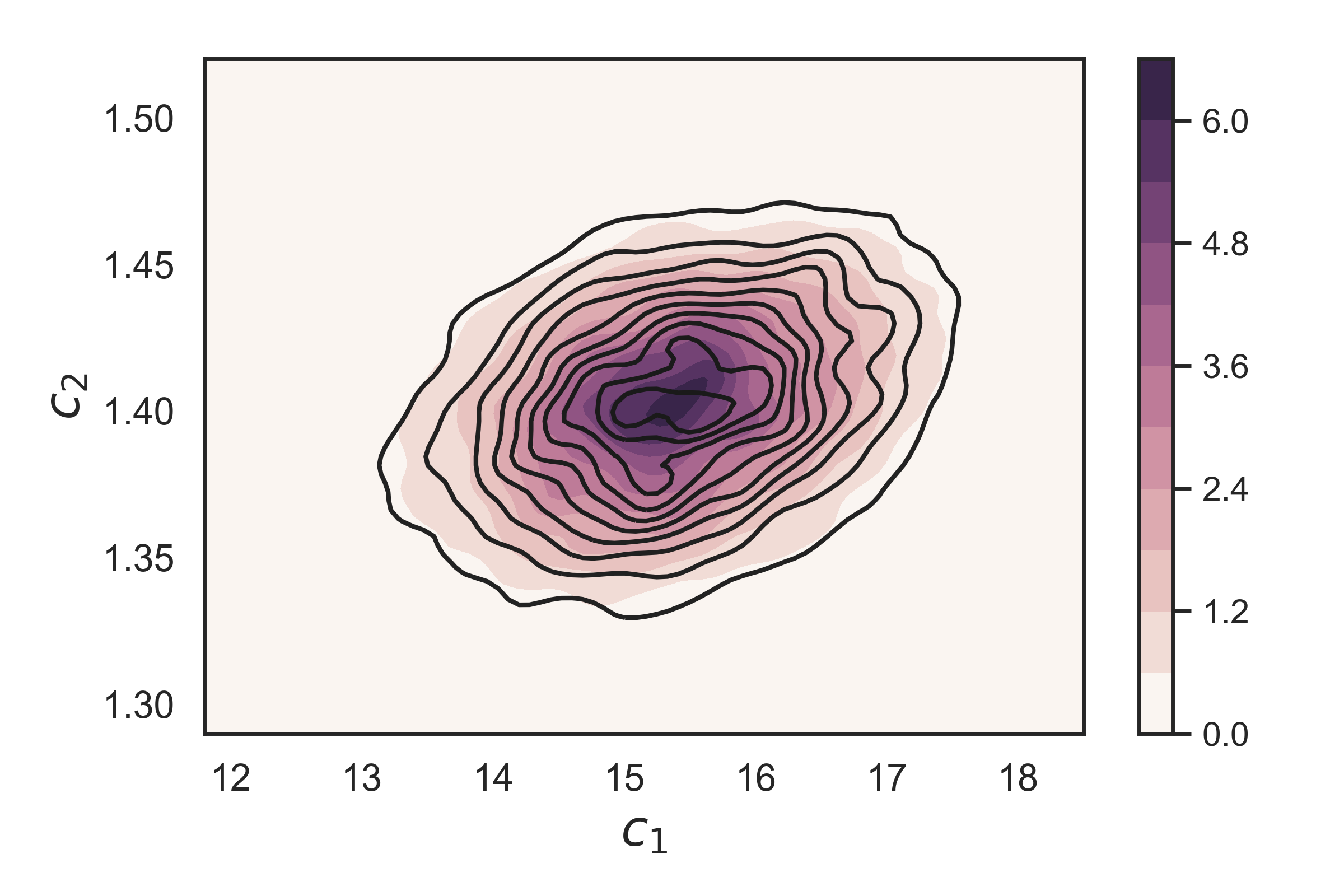}
			\caption{}
		\end{subfigure}
		\captionsetup{}
		\caption{Estimated joint posterior distributions: (a) a comparison between MH-FD (shaded area) and MH-PINN (dashed line) results; (b) a comparison between MH-FD (shaded area) and MH-APINN (solid line) results.} 
		\label{C8_fig_density_joint}
	\end{center}
\end{figure}

Figures \ref{C8_fig_density_marginal}, \ref{C8_fig_density_joint} show the results for the estimated marginal and joint posterior distributions for parameters $c_1$ and $c_2$, using the MH-FD, MH-PINN-UQ, and MH-APINN (with $\epsilon_t$ set to 0.025) methods. The acceptance rate for MH-FD, MH-PINN-UQ, and MH-APINN samplers are, respectively, 25.20\%, 26.82\%, and 25.43\%. From these two figures, it is evident that unlike the MH-PINN-UQ results, the MH-APINN results are in good agreement with those of MH-FD. Moreover, Table \ref{C8_table_time} shows the execution time for the MH-FD, MH-PINN-UQ, and MH-APINN methods. 

\begin{table}[h]
	\begin{center}
		\begin{tabular}{|c||c|c|c|}
			\hline
			Sampling method          & MH-FD & MH-PINN-UQ & MH-APINN \\ \hline
			Execution time (minutes) & 1,507 & 35         & 43       \\ \hline
		\end{tabular}
		\caption{Execution time for the MH-FD, MH-PINN-UQ, and MH-APINN methods, for solving the parameter estimation problem defined in Equation \ref{C8_Eq_PDE}.}
		\label{C8_table_time}
	\end{center}
\end{table}

\begin{figure}
	\begin{center}
		\includegraphics[width=0.7\linewidth]{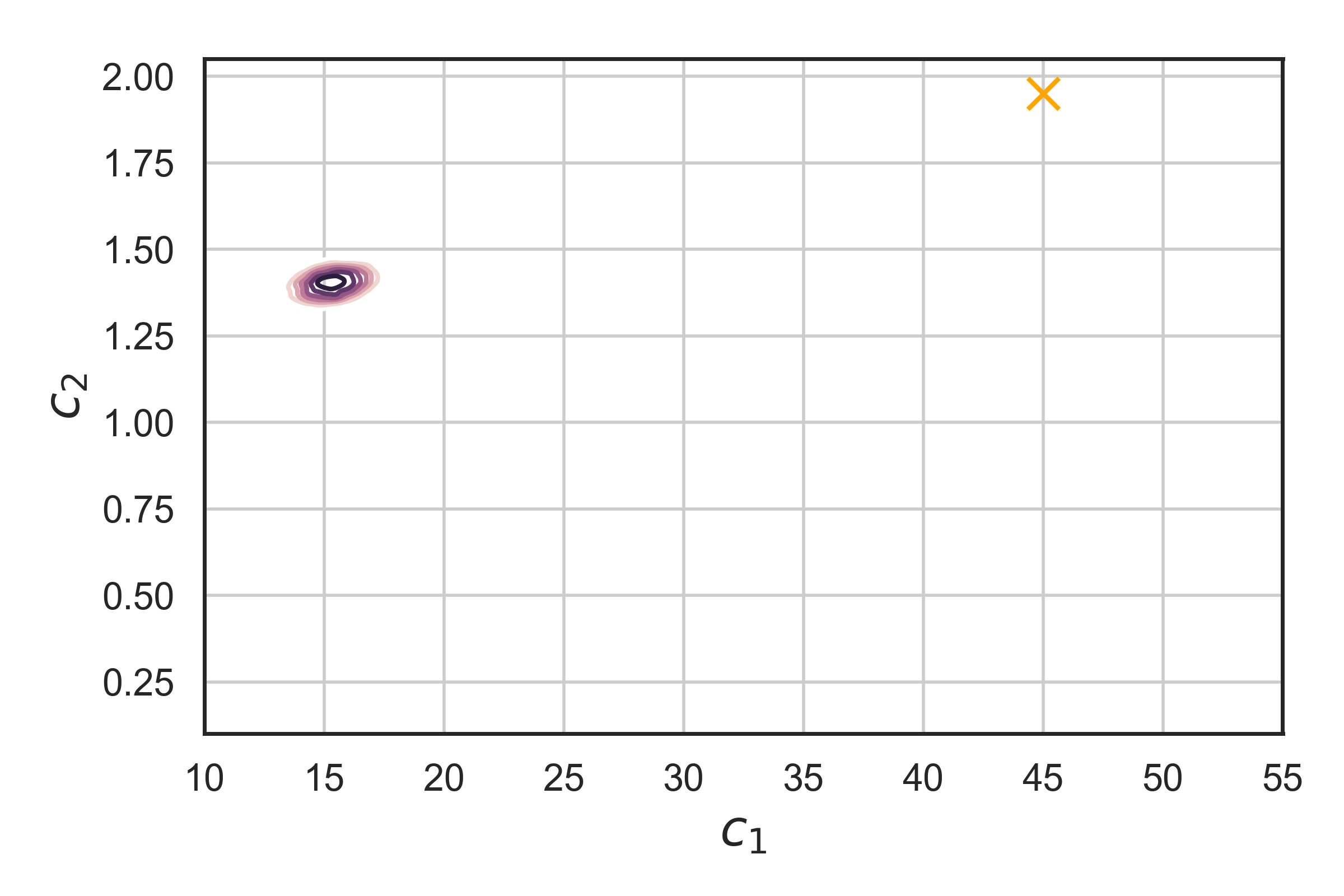}
		\caption{Posterior distribution of parameters $c_1$ and $c_2$ in a quarter of support of the prior distribution. The cross symbol represents the initial state of our MH samplers.} 
		\label{C8_fig_posterior_vs_prior}
	\end{center}
\end{figure}

As stated earlier, the posterior distribution is usually concentrated on a small portion of prior distribution, and thus, it is natural to train a approximate model to $f$ that is fine-tuned in a region where posterior resides. To verify this, Figure \ref{C8_fig_posterior_vs_prior} is depicted, showing the posterior distribution of parameters $c_1$ and $c_2$ in only a quarter of the prior support. The cross symbol represents the initial state of our MH samplers.

\begin{figure}
	\begin{center}
		\begin{subfigure}{0.6 \linewidth}
			\includegraphics[width=.99\linewidth]{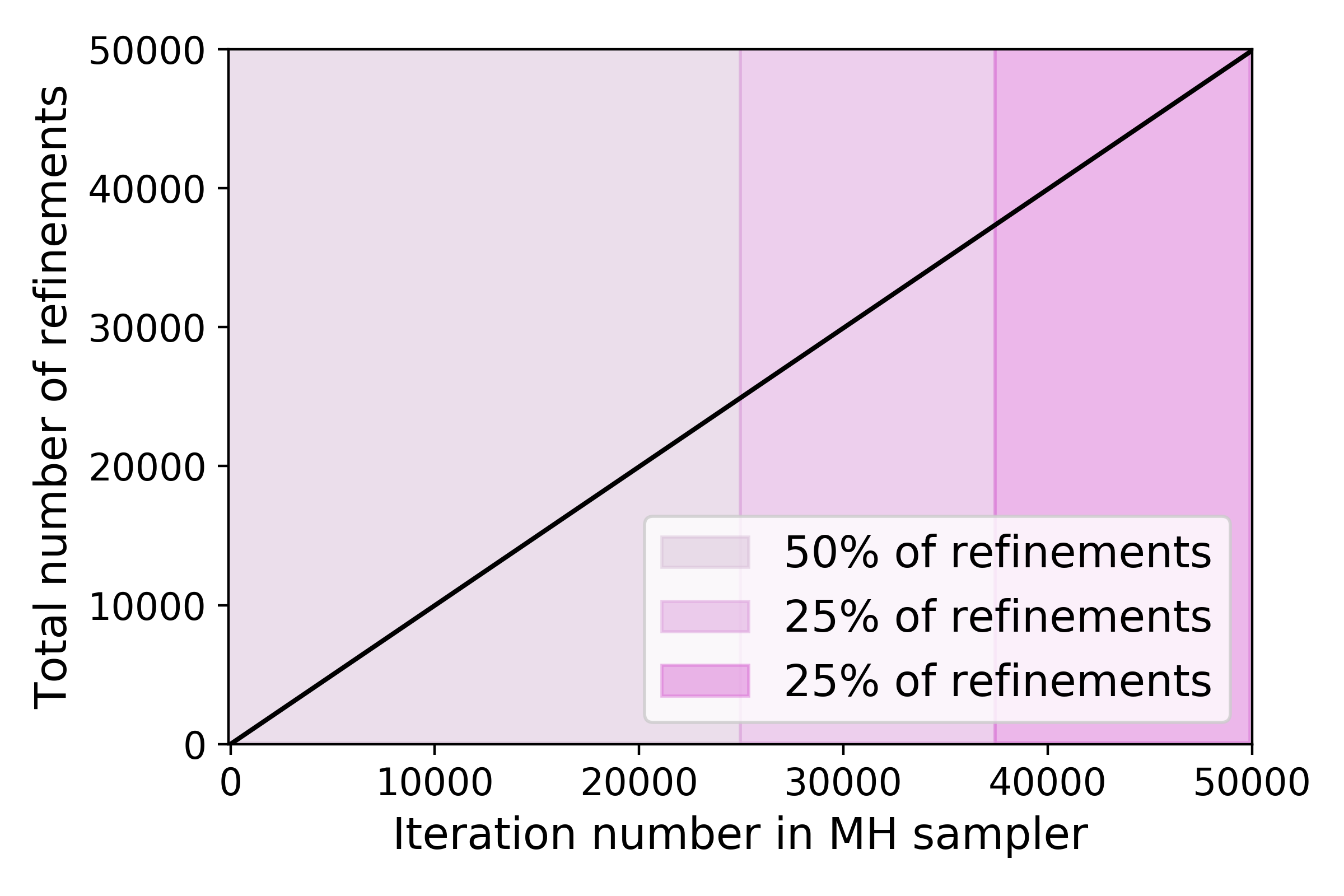}
			\caption{}
			\centering	
		\end{subfigure}
		\quad
		\begin{subfigure}{0.6 \linewidth}
			\includegraphics[width=.99\linewidth]{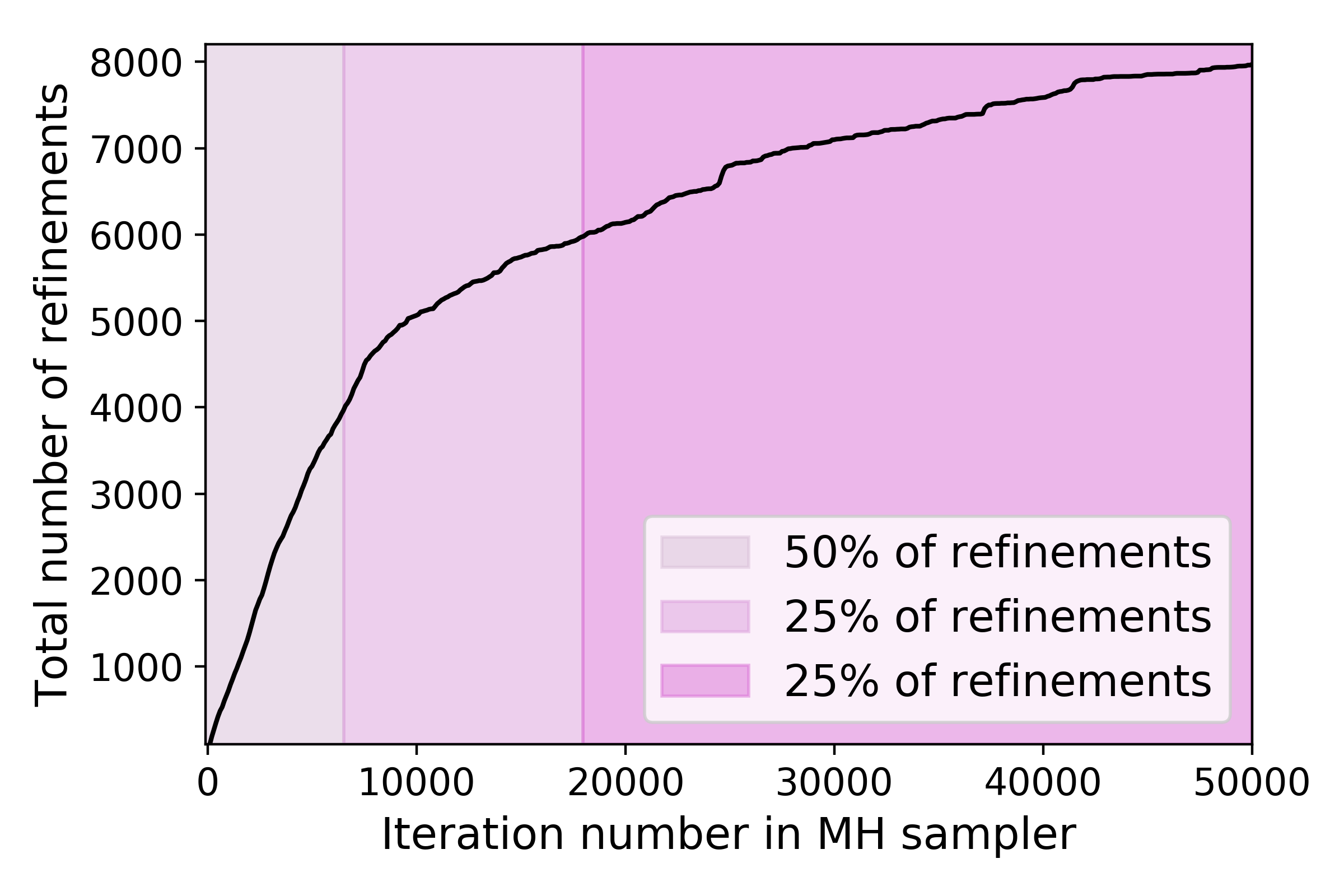}
			\caption{}
		\end{subfigure}
		\\
		\begin{subfigure}{0.6 \linewidth}
			\includegraphics[width=.99\linewidth]{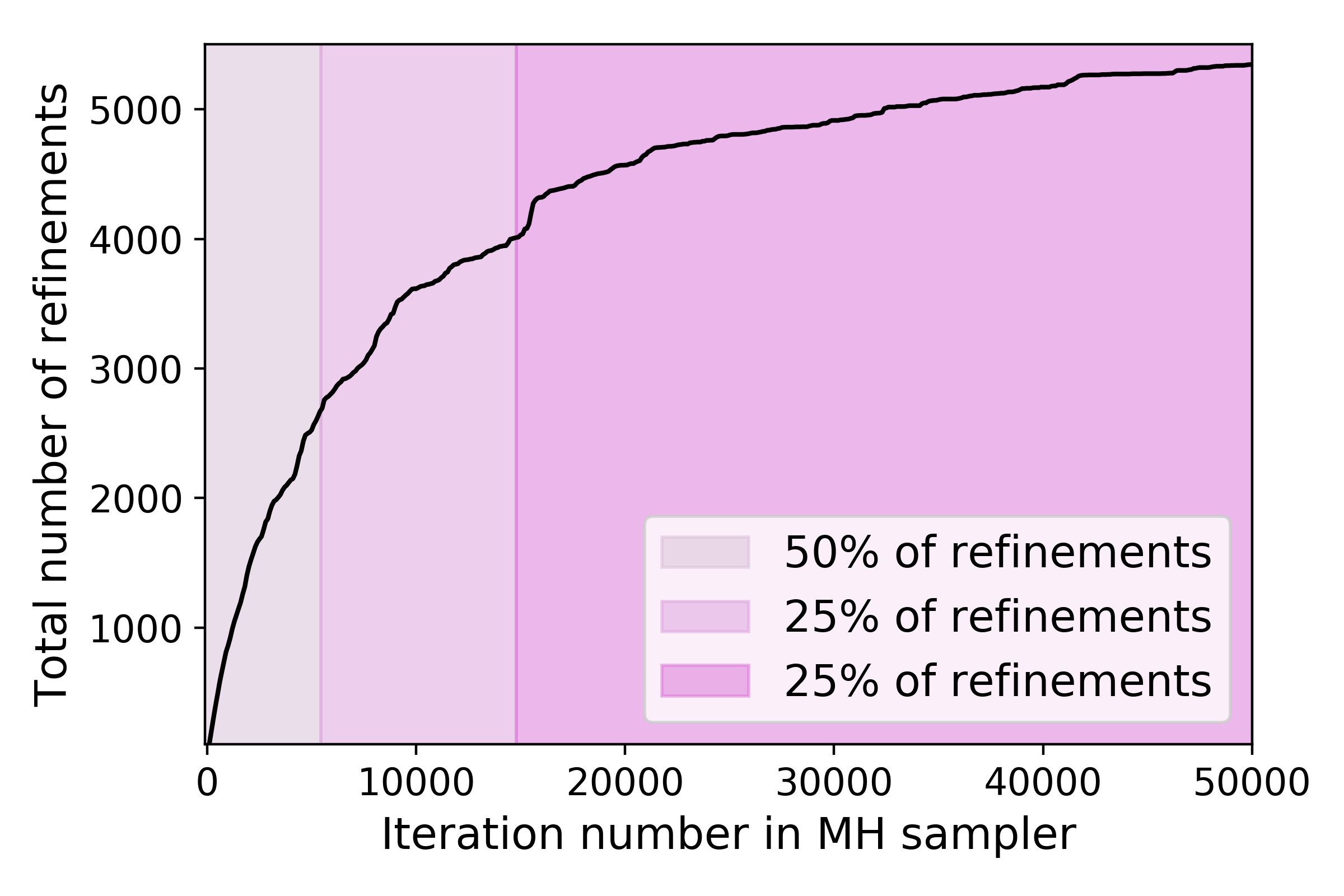}
			\caption{}
		\end{subfigure}
		\captionsetup{}
		\caption{the accumulated number of online surrogate refinement versus the number of MH iterations, for three refinement options: (a) local surrogate is refined but none of these refinements are reflected in the global surrogate; (b) local surrogate is refined and all of these refinements are reflected in the global surrogate; and (c) local surrogate is refined and only the first iteration of model parameter update is  reflected in the global surrogate, as implemented in the APINN algorithm.} 
		\label{C8_fig_refinement_rate}
	\end{center}
\end{figure}

Figure \ref{C8_fig_refinement_rate} shows the cumulative number of online surrogate refinement versus the number of MH iterations, for three refinement options: (a) local surrogate is refined but none of these refinements are reflected in the global surrogate; (b) local surrogate is refined and all of these refinements are reflected in the global surrogate; and (c) local surrogate is refined and only the first iteration of model parameter update is  reflected in the global surrogate, as implemented in the APINN algorithm. The slope of this cure represents the rate for which the surrogate  is refined. Evidently, for APINN, the slope of the curve for the initial iterations of the MH sampler is relatively high, and gradually decreases as the surrogate is refined based on the samples collected from the posterior distribution. 

\begin{figure}
	\begin{center}
		\begin{subfigure}{0.58 \linewidth}
			\includegraphics[width=.99\linewidth]{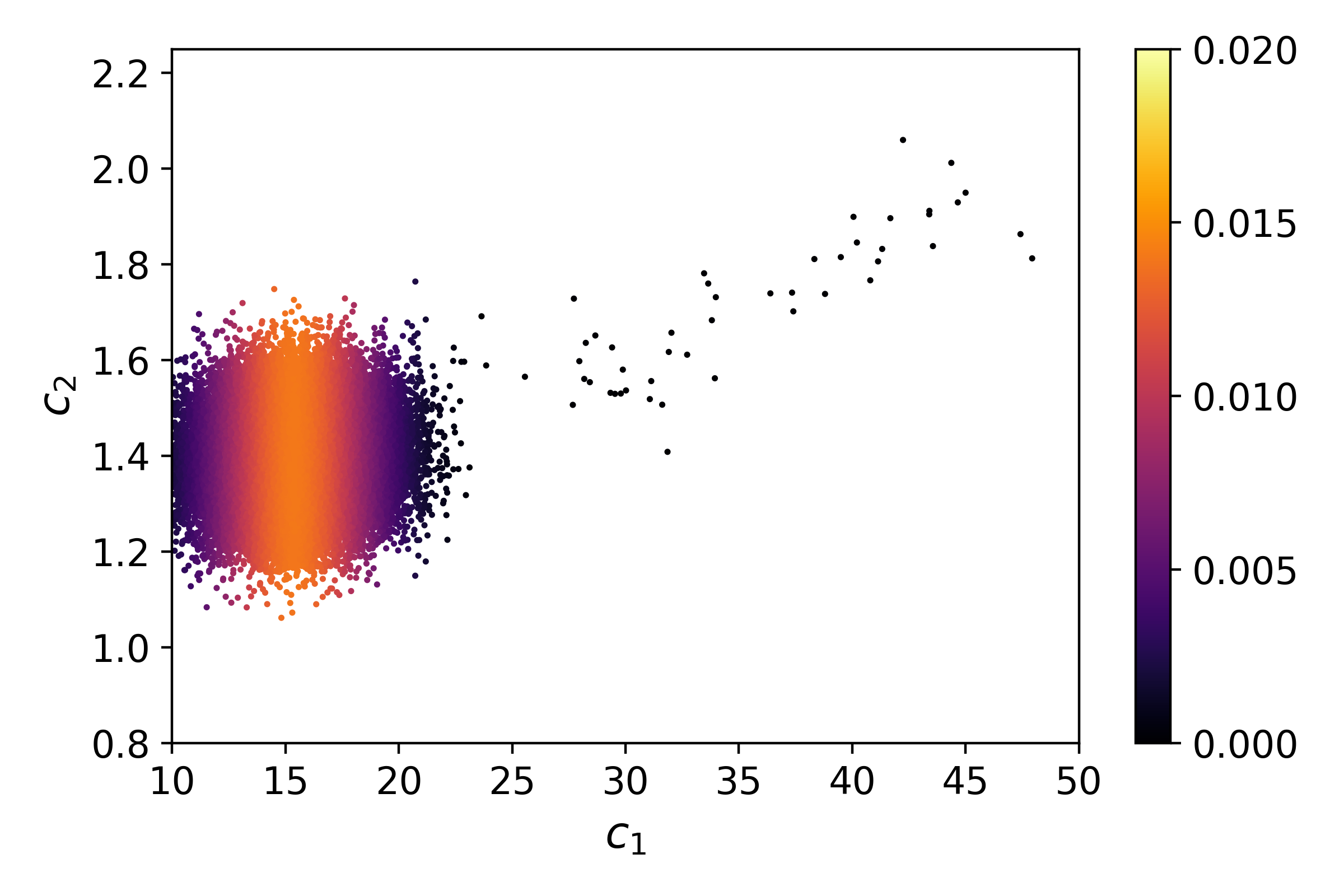}
			\caption{}
			\centering	
		\end{subfigure}
		\quad
		\begin{subfigure}{0.58\linewidth}
			\includegraphics[width=.99\linewidth]{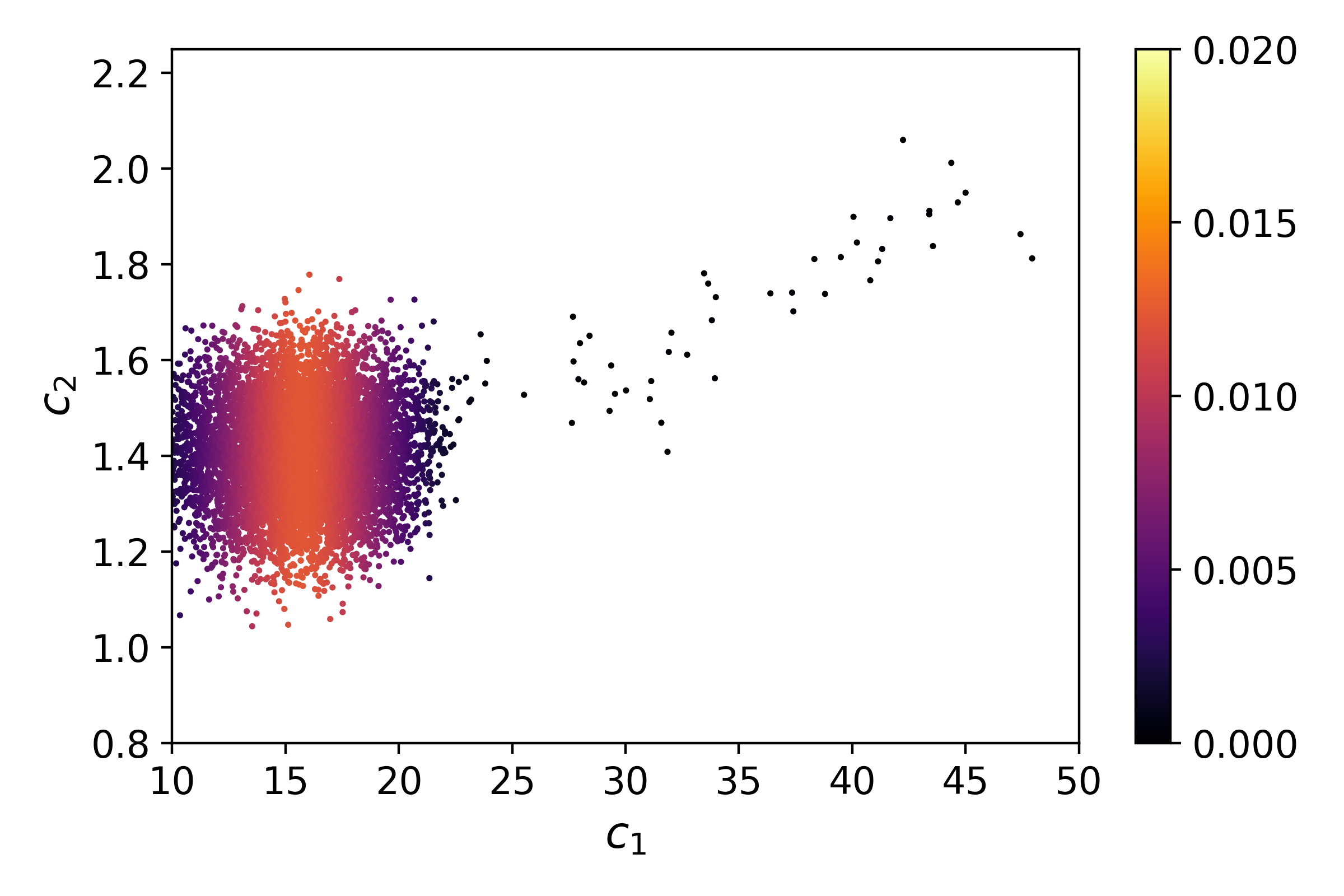}
			\caption{}
		\end{subfigure}
		\\
		\begin{subfigure}{0.58 \linewidth}
			\includegraphics[width=.99\linewidth]{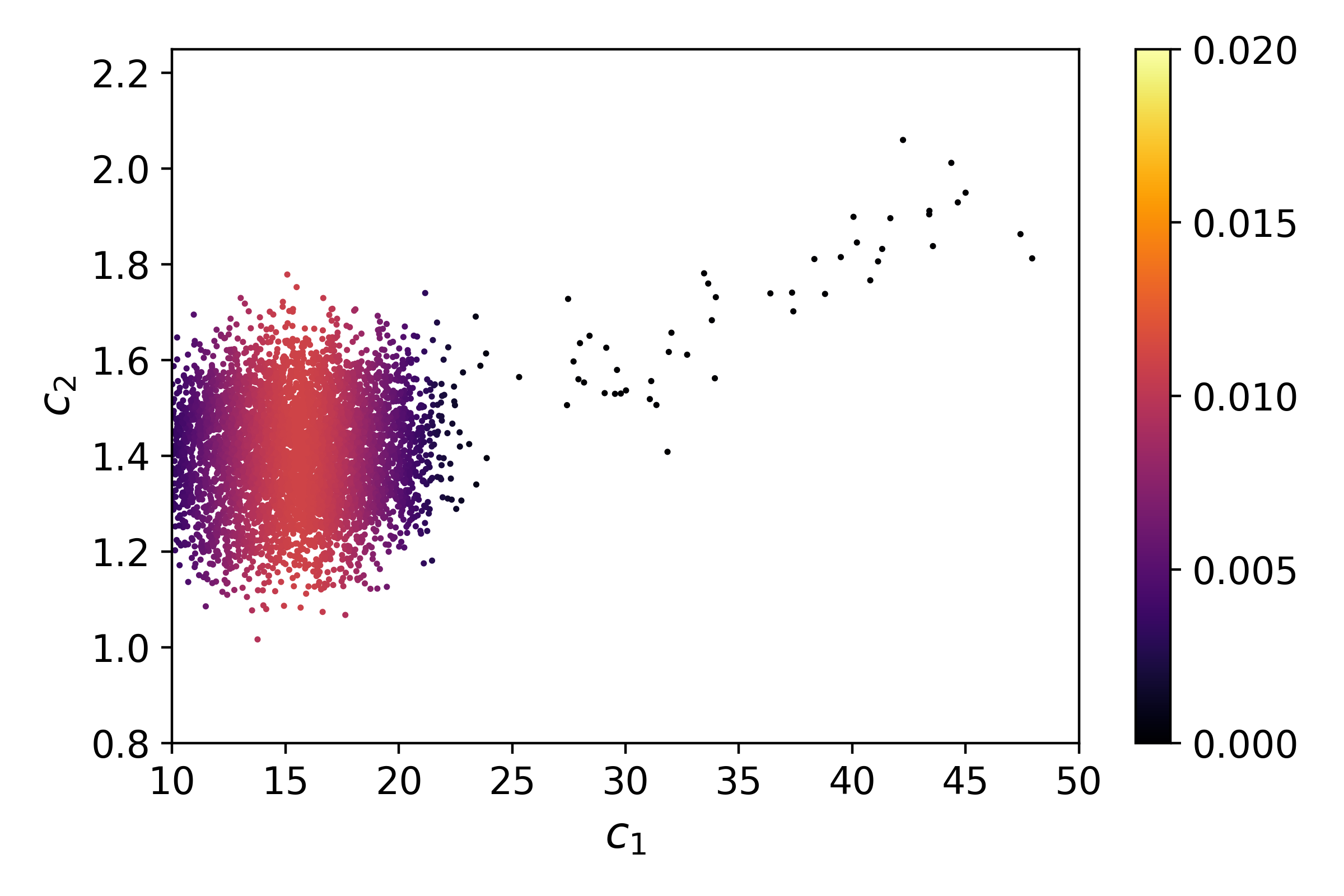}
			\caption{}
		\end{subfigure}
		\captionsetup{}
		\caption{The MH samples for which a local surrogate refinement is executed, for three different refinement options: (a) local surrogate is refined but none of these refinements are reflected in the global surrogate; (b) local surrogate is refined and all of these refinements are reflected in the global surrogate; and (c) local surrogate is refined and only the first iteration of model parameter update is  reflected in the global surrogate, as implemented in the APINN algorithm.} 
		\label{C8_fig_sample_density}
	\end{center}
\end{figure}

For each of the three refinement options for the global surrogate, Figure \ref{C8_fig_sample_density} depicts the MH samples for which a local surrogate refinement is executed. The total number of MH samples for which a local refinement is executed is 49,902, 7,970, and 5,347, respectively, for the three outlined options. The total number of training iterations for local refinements is 4,976,715, 46,690, and 37,847, respectively, for the three options. It is evident that the proposed online training for APINNs shows superior performance in terms of efficiency compared to the other two online training options, based on the total number of refinements and the total number of refinement iterations.

\section{Concluding Remarks} 

In many parameter estimation problems in engineering systems, the forward model $f$ consists of solving a PDE. In this case, computing the forward model $f$ in an MCMC simulation can be computationally expensive or even intractable, as usually MCMC samplers require thousands of millions of iterations to provide converged posterior distributions, and the model $f$ needs to be computed at each and every of these iterations. Constructing a global approximating surrogate over the entire prior support it computationally inefficient as the posterior density can reside only on a small fraction of the prior support. To alleviate this computational limitation, we presented a novel adaptive method, called APINN, for efficient MCMC-based parameter estimation. The proposed method consists of: (1) constructing an offline surrogate model as an approximation to the forward model $f$; and (2) refining this approximate model on the fly using the MCMC samples generated from the posterior distribution. An important feature of the proposed APINN method is that for each likelihood evaluation, it can always bound the approximation error to be less than a user-defined residual error threshold to ensure the accuracy of the posterior estimation. The promising performance of the proposed APINN method for MCMC was illustrated through a parameter estimation example for a system governed by the Poisson equation. Moreover, the efficiency of the APINN online refinement scheme was illustrated in comparison with two other competing schemes.


\bibliography{bibfile}

\end{document}